\documentclass[sigconf,nonacm]{acmart}
\AtBeginDocument{%
  }

\usepackage{latexsym}
\usepackage[utf8]{inputenc}
\usepackage{bbm}
\usepackage{microtype}

\usepackage{graphicx}

\usepackage{tabularx} 
\usepackage{longtable}

\usepackage{booktabs}
\usepackage{multirow}
\usepackage{graphicx}

\usepackage{algorithm}
\usepackage{algorithmic}

\usepackage{newfloat}
\usepackage{listings}

\usepackage{subfigure}
\usepackage{mathrsfs}
\usepackage{lineno}
\usepackage[table]{xcolor}
\usepackage{color, soul}
\usepackage[inline]{enumitem}
\usepackage{acronym}

\usepackage{tikz}
\usetikzlibrary{calc,shapes.callouts}
\usepackage[most]{tcolorbox}
\tcbuselibrary{listings,breakable}

\usepackage[skip=2pt]{caption}

\usepackage{amsmath}
\usepackage{arydshln}
\usepackage{fontawesome}

\usepackage{changepage}
\usepackage{mdframed}
\usepackage{pifont}
\usepackage{makecell}
\usepackage[most]{tcolorbox}

\AtBeginDocument{%
  }

\acrodef{IR}{information retrieval}
\acrodef{LLM}{large language model}
\acrodef{RLAIF}{reinforcement learning from AI feedback}
\acrodef{RLHF}{reinforcement learning from human feedback}
\acrodef{DPO}{direct preference optimization}
\acrodef{PRO}{preference ranking optimization}

\acrodef{Fine-grained PA}{Fine-grained Preference Alignment}
\acrodef{NLP}{natural language processing}
\acrodef{KnowTuning}{knowledge-aware fine-tuning}
\acrodef{SFT}{supervised fine-tuning}
\acrodef{SPO}{Subject-Predicate-Object}
\acrodef{QA}{question answering}
\acrodef{PPO}{Proximal Policy Optimization}
\acrodef{KILT}{knowledge intensive language tasks}
\acrodef{SACD}{self-adaptive cognitive debiasing}

\definecolor{darkgreen}{rgb}{0.0, 0.5, 0.0}

\definecolor{darkblue}{HTML}{2E75B6}

\newcommand{\header}[1]{\vspace{1.5mm}\noindent\textbf{#1}.}

\setcopyright{cc}
\copyrightyear{2026}
\acmYear{2026}
\acmConference[KDD '26]{The 32th International ACM SIGKDD Conference on Knowledge Discoveryand Data Mining}{August 9–13, 2026}{Jeju, Korea}
\acmISBN{}
\acmDOI{}


\author{%
Yougang Lyu\textsuperscript{1} \quad Xi Zhang\textsuperscript{1} \quad Yuyue Zhao\textsuperscript{1} \quad Konstantinos Papakostas\textsuperscript{1} \quad Xinhao Yi\textsuperscript{1} \\
Shuyu Guo\textsuperscript{1} \quad Wenxiang Hu\textsuperscript{1} \quad Jan Piotrowski\textsuperscript{1} \quad Jakub Kaliski\textsuperscript{1} \quad Jacopo Urbani\textsuperscript{1,2} \\
Zaiqiao Meng\textsuperscript{1} \quad Lun Zhou\textsuperscript{1} \quad Zhaochun Ren\textsuperscript{3} \quad Xiaohui Yan\textsuperscript{1}%
}
\authornote{Corresponding author.}
\affiliation{%
  \institution{\textsuperscript{1}Huawei Technologies Co., Ltd. \quad \textsuperscript{2}Vrije Universiteit Amsterdam \quad \textsuperscript{3}Leiden University \\
    youganglyu@gmail.com, yanxiaohui2@huawei.com}
  \country{}
}

\renewcommand{\shortauthors}{Yougang Lyu et al.}

\begin{document}

\title[EvoScientist: Towards Multi-Agent Evolving AI Scientists for \\End-to-End Scientific Discovery]{
EvoScientist: Towards Multi-Agent Evolving AI Scientists for \\
End-to-End Scientific Discovery
}

\begin{abstract}
The increasing adoption of Large Language Models (LLMs) has enabled AI scientists to perform increasingly complex end-to-end scientific discovery tasks. Such tasks required the coordination of specialized roles, including idea generation and experimental execution.  Despite this complexity, most state-of-the-art AI scientist systems rely on static, hand-designed pipelines and fail to adapt their idea- or code-generation strategies based on accumulated interaction histories. As a result, these systems systematically overlook promising research directions, repeat previously failed experiments, and pursue infeasible ideas. %
To address this limitation, we introduce \textbf{EvoScientist}, an evolving multi-agent AI scientist framework that continuously improves its research strategies through persistent memory and self-evolution. EvoScientist comprises three specialized agents: a Researcher Agent (\textit{RA}) responsible for scientific idea generation, an Engineer Agent (\textit{EA}) responsible for experiment implementation and execution, and an Evolution Manager Agent (\textit{EMA}) that distills insights from prior agent interactions into reusable knowledge. Specifically, EvoScientist contains two persistent memory modules: (i) an ideation memory, which summarizes feasible research directions from top-ranked ideas while recording previously unsuccessful directions identified during idea validation; and (ii) an experimentation memory, which captures effective data processing and model training strategies derived from code search trajectories and best-performing implementations. These memory modules enable the \textit{RA} and \textit{EA} to retrieve relevant prior strategies, thereby improving idea quality and increasing code execution success rates over time. Experiments show that EvoScientist outperforms \textbf{7} open-source and commercial state-of-the-art systems in scientific idea generation, achieving higher performance in terms of novelty, feasibility, relevance, and clarity through automatic and human evaluation. Furthermore, EvoScientist substantially improves code execution success rates through multi-agent evolution, demonstrating the effectiveness of persistent memory for end-to-end scientific discovery.

\smallskip
\noindent\faGithub~\textbf{Code:} \url{https://github.com/EvoScientist/EvoScientist}

\end{abstract}

\maketitle

\acresetall


\section{Introduction}
\label{sec:introduction}
Scientific discovery progresses through a recurring cycle of observation, hypothesis formation, experimental testing, and application, in which researchers systematically explore existing knowledge, synthesize new ideas, and refine their understanding through empirical feedback~\citep{langley1987scientific,klahr1999studies,popper2005logic}. Traditionally, this process has been driven by expert scientists who read extensive literature, formulate hypotheses, and validate them through rigorous experimentation, gradually accumulating experience into scientific expertise~\citep{klahr2000exploring,kuhn1970structure,platt1964strong}. However, the vast and rapidly expanding space of possible concepts, mechanisms, and experimental conditions fundamentally limits how quickly humans can explore, evaluate, and verify new ideas~\citep{gridach2025agenticaiscientificdiscovery,reddy2024scientificdiscoverygenerativeai}. This challenge is further amplified by the explosive growth of scientific publications, making it increasingly difficult and time-consuming to keep up with the literature, generate novel yet feasible ideas, and execute validation experiments~\citep{weng2025deepscientistadvancingfrontierpushingscientific,shao2025omniscientistcoevolvingecosystemhuman}.

To substantially accelerate research, \textit{AI-driven scientific discovery} has progressed from applying Large Language Models (LLMs) to isolated sub-tasks to building agentic systems that support coordinated scientific reasoning and action across the discovery process~\citep{chen2025ai4research}. One line of work focuses on early-stage idea generation, where LLMs and multi-agent collaboration are used to propose, critique, and iteratively refine hypotheses~\citep{si2024can,gottweis2025towards,li2024learning,gao2025graph,qi2024large,o2025sparks,azher2025futuregen,sanyal2025spark,su2025many}. Representative work such as Virtual Scientist (VirSci)~\citep{su2025many} and Co-Scientist~\citep{gottweis2025towards} organizes multiple agents to simulate collaborative scientific ideation through proposal, critique, and refinement~\citep{su2025many}. 
In parallel, a second line of work develops end-to-end AI scientist systems that automate the workflow from ideation and literature review to experiment implementation, analysis~\citep{lu2024ai,yamada2025ai,zochi2025,schmidgall2025agent,weng2025deepscientist,shao2025omniscientist,team2025novelseek,tang2025ai}. Examples include AI Scientist-v2~\citep{yamada2025ai}, which employs agentic tree search to improve end-to-end research trajectories, 
AI-Researcher~\citep{tang2025ai}, which orchestrates structured collaboration across the full research pipeline, and InternAgent~\citep{team2025novelseek}, which incorporates human expert feedback into the agent workflow.

Although these systems demonstrate encouraging progress, they largely treat end-to-end scientific discovery as a static execution pipeline. Agent roles, decision strategies, and interaction patterns are typically fixed after deployment, and accumulated outcomes and failures are rarely distilled into reusable experience.
As a result, such system may repeatedly explore known failure patterns, overlook promising research directions, or invest substantial resources in infeasible ideas.
These limitations highlight a missing capability in existing AI scientist systems: \textit{the ability to learn from accumulated outcomes and failures and to continuously improve both idea generation and experiment execution over time.}
This motivates the formulation of \textit{multi-agent evolution} as
a core requirement for end-to-end scientific discovery, where interaction histories are treated as a first-class resource rather than discarded execution traces.
Accordingly, we study the following research question:
{
\setlength{\leftmargini}{5pt}
\setlength{\rightmargin}{5pt}
\setlength{\topsep}{4pt}
\setlength{\partopsep}{0pt}
\begin{quote} 
\textit{How can we formulate end-to-end scientific discovery as a learning problem in which multi-agent systems evolve their idea-generation and-code generation by learning from prior successes and failures?}
\end{quote}
}

To answer this question, we propose \textbf{EvoScientist}, a multi-agent evolution framework designed to solve the above end-to-end scientific discovery problem. EvoScientist decomposes scientific discovery into three specialised agents: a Researcher Agent (\textit{RA}) that generates scientific ideas and research proposals, an Engineer Agent (\textit{EA}) that executes experiments and produces code and analysis, and an Evolution Manager Agent (\textit{EMA}) that distills interaction histories into persistent memories to guide future decision-making. Specifically, EvoScientist implements multi-agent evolution through two memory modules: (i) an \textit{ideation memory}, which summarizes high-quality research directions from top-ranked ideas while recording directions that failed during idea validation; and (ii) an \textit{experimentation memory}, which captures effective data processing and model training strategies derived from code search trajectories and the best-performing implementations. For each new task, the \textit{RA} and \textit{EA} retrieve relevant strategies from these memories and append them to their prompts, enabling continuous improvement in idea quality and code execution success rates over time.

We conduct experiments on scientific idea generation, code generation, and end-to-end scientific discovery. EvoScientist outperforms \textbf{7} open-source and commercial baselines in idea generation quality (measured in terms of novelty, feasibility, relevance, and clarity) under both automatic and human evaluation, and achieves higher code execution success rates through multi-agent evolution. In an end-to-end evaluation, all six full papers generated by EvoScientist were accepted to ICAIS 2025~\citep{icais2025} (\textit{AI Scientist Track}), and two received major awards (the \textit{Best Paper Award} and the \textit{AI Reviewer's Appraisal Award}). 
In summary, our main contributions are:
\begin{itemize}[leftmargin=*,topsep=4pt, itemsep=4pt]

    \item[\ding{182}] We propose EvoScientist, a self-evolving multi-agent system with three specialized agents and two persistent memory modules, aiming to improve both the quality of generated research ideas and the reliability of code generation and execution. 

    \item[\ding{183}] We introduce three multi-agent self-evolution mechanisms, namely idea direction evolution, idea validation evolution, and experiment strategy evolution, that enable EvoScientist to learn from accumulated outcomes and failures and to continuously improve both idea generation and experiment execution over time.

    \item[\ding{184}] We provide empirical evidence that EvoScientist generates higher-quality ideas and achieves higher code execution success rates compared to strong open-source and commercial baselines.
\end{itemize}

\section{Related Work}
\label{sec:related_work}
\subsection{AI Agents for Scientific Discovery}
The application of AI to scientific discovery has rapidly progressed from assisting with discrete research tasks to integrated, autonomous agents capable of managing large capable  of managing increasingly large portions of the research lifecycle~\citep{chen2025ai4research}. Early work established that LLMs can serve as effective tools for specific sub-tasks, particularly early-stage ideation. A growing body of studies has shown that LLMs can propose novel and high-quality research ideas that are competitive with those of human experts, highlighting their potential as creative aids in scientific ideation~\citep{si2024can,li2024learning,gao2025graph,qi2024large}. 
Systems such as HypoGen~\citep{o2025sparks} and Futuregen~\citep{azher2025futuregen} analyze scientific literature to identify knowladge gaps and propose novel research questions, while other approaches, including Spark~\citep{sanyal2025spark} and ResearchBench~\citep{liu2025researchbench}, demonstrate that LLMs can generate feasible and creative research ideas by leveraging preteainedknowledge and retrieved evidence from the literature. Building on this line of work, Virtual Scientist (VirSci)~\citep{su2025many} employs multi-agent collaboration to simulate scientific teamwork for proposing, evaluating, and refining ideas, illustrating how coordinated agent architectures can enhance early-stage ideation.

More recently, the field has shifted towards developing end-to-end scientific discovery agents that aim to automate the scientific workflow across multiple stages, including ideation and literature review, experimental design, code implementation, data analysis, and even manuscript preparation~\citep{lu2024ai,yamada2025ai,zochi2025,schmidgall2025agent}. A seminal example is The AI Scientist~\citep{lu2024ai}, which demonstrated a full pipeline from idea generation to manuscript writing. Its successor, The AI Scientist-v2~\citep{yamada2025ai}, further improved end-to-end performance by incorporating agentic tree search to explore alternative research trajectories. Other systems investigate different facets of autonomous research using multi-agent architectures with specialized roles (e.g., proposers, experimenters, and critics) to simulate collaborative scientific processes~\citep{schmidgall2025agentrxiv,schmidgall2025agent}. 
For instance, AgentArxiv~\citep{schmidgall2025agentrxiv} and AgentLab~\citep{schmidgall2025agent} explicitly model iterative collaboration among agents, while AI co-scientist~\citep{gottweis2025towards} adopts a ``generate, debate, and refine'' paradigm to tackle complex biomedical research problems. AI-Researcher~\citep{tang2025ai} orchestrates a structured multi-agent workflow spanning literature analysis, experiment execution, and manuscript preparation, and InternAgent~\citep{team2025novelseek} incorporates scalable human expert feedback into the agent loop. Beyond general-purpose research automation, some systems explore long-horizon or goal-driven discovery settings; for example, DeepScientist~\citep{weng2025deepscientist} formulates scientific discovery as sequential experimental optimization over extended timelines, while OmniScientist~\citep{shao2025omniscientist} models a broader social and collaborative ecosystem of human science, such as peer review and knowledge sharing.

Despite these advances, improvements in existing AI scientist systems are typically confined to \textit{within-run} exploration mechanisms, such as tree search, debate, or Bayesian optimization. Agent roles and decision policies are often pre-specified and remain largely unchanged across tasks, and interaction outcomes and failures are rarely distilled into persistent, reusable experience that can inform future ideation and experiment execution.
Consequently, such systems may repeatedly revisit known failure patterns, overlook promising research directions, or invest substantial resources in experimentally infeasible ideas. This limitation motivates AI scientist systems that not only execute end-to-end research pipelines, but also support \textit{multi-agent evolution} by systematically learning from accumulated interaction histories.

\vspace{-10pt}

\subsection{Self-Evolving Agents}
While powerful, most contemporary LLM-based agents rely on fixed, pre-specified policies and do not reliably adapt their core decision-making strategies in response to new information or failures. This limitation has become a critical bottleneck, particularly in dynamic and long-horizon environments, motivating growing interest in \textit{self-evolving agents} that can continually learn from their experiences~\citep{fang2025comprehensive, gao2025survey}. The primary advantage of such agents is their ability to adaptively reason and act over time, leading to improved robustness and generalization across tasks.

The development of self-evolving agents is driven by mechanisms that enable the modification of agent behavior based on experience.
Among the most prominent are memory systems, which allow agents to store and retrieve, and consolidate information from past interactions and outcomes~\citep{chhikara2025mem0,wang2024agent,zhao2024expel}, and adaptive tool-use frameworks, which expand agent capabilities by enabling the autonomous creation, refinement, and management of tools~\citep{qiu2025alita,qu2024exploration,wang2023voyager}. 
Agent evolution is further supported by learning paradigms such as reward-based learning from feedback signals~\citep{shinn2023reflexion}, imitation-based learning from expert demonstrations~\citep{zelikman2022star}, and population-based or evolutionary methods inspired by biological evolution~\citep{zhang2025darwin}. 
These approaches have demonstrated promising results across a range of applications domains, including coding~\citep{robeyns2025self,wang2024rlcoder}, education~\citep{liu2025one}, and healthcare~\cite{almansoori2025medagentsim}, where agents can progressively tailor their behavior to specific tasks and user needs. 

Despite this progress, existing self-evolving agents are predominantly evaluated on single-stage or narrowly scoped tasks, and their evolution mechanisms are rarely designed to support the multi-stage requirements of end-to-end scientific discovery. 
In particular, they have not been shown to evolve \textit{both} ideation and experiment-execution strategies under a unified objective that spans idea generation, validation, and experimental implementation. Our work addresses this gap by instantiating self-evolving agents in the context of end-to-end scientific discovery, where multi-agent systems learn from accumulated interaction histories to improve performance across the full discovery pipeline.
\begin{figure*}[htbp]
  \centering
\includegraphics[width=0.9\textwidth]{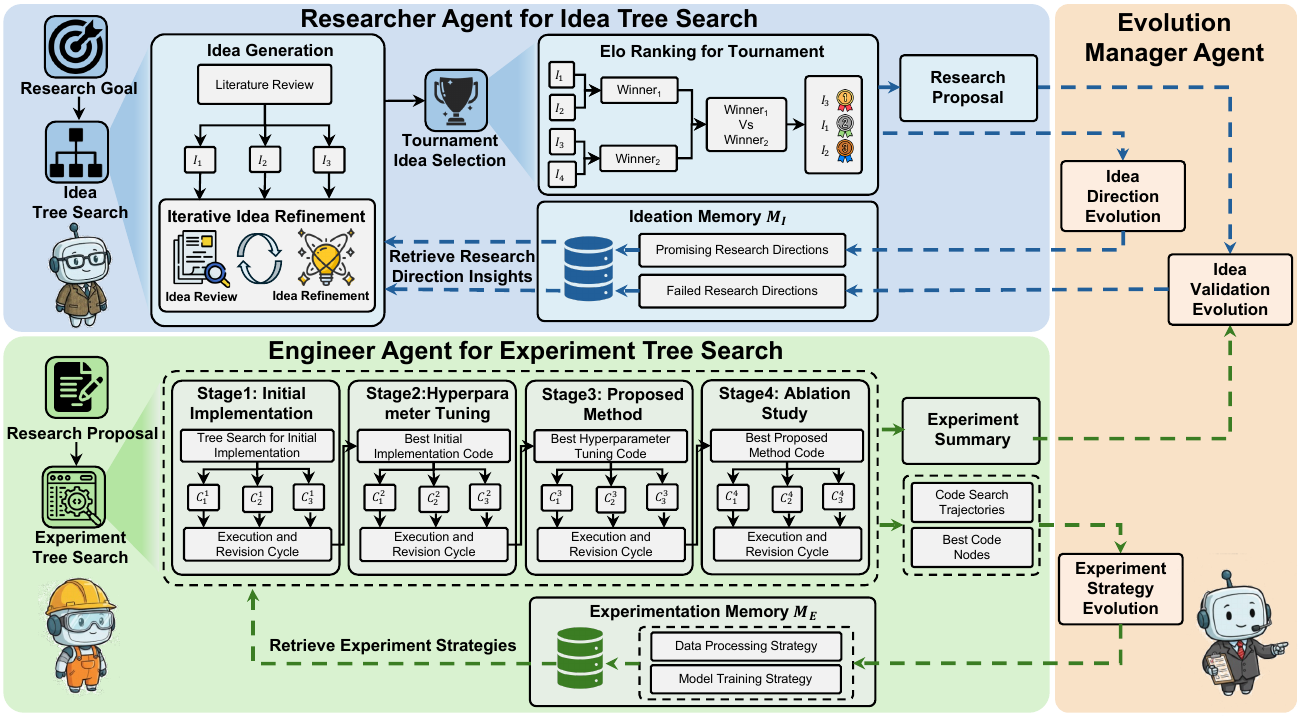}
\vspace*{-1mm}
\caption{Overview of EvoScientist, a self-evolving multi-agent system for end-to-end scientific discovery. EvoScientist consists of a researcher agent (RA), an engineer agent (EA), and an evolution manager agent (EMA). The EMA distills interaction histories into two persistent memories, an ideation memory $M_I$ and an experimentation memory $M_E$, which are retrieved by the RA and EA to enable continuous improvement in idea quality and execution success rates across tasks.}
\vspace*{-4.5mm}
\label{fig:framework}
\end{figure*}

\section{Method}
\label{sec:method}
In this section, we detail the EvoScientist method. First, we formulate our research problem. Then, we introduce the framework of EvoScientist. Next, we introduce the research agent for idea tree search and the engineer agent for experiment tree search. Finally, the evolution manager agent for multi-agent evolution is explained.

\subsection{Problem Formulation}
Following~\citet{weng2025deepscientist,tang2025ai,shao2025omniscientist}, we define end-to-end scientific discovery as a goal-driven and verifiable pipeline that transforms a user goal $G$ into a proposal and executable experiments. The key challenge is to jointly improve idea quality and execution reliability by learning from outcomes and failures accumulated across tasks. Specifically, the pipeline proceeds in two stages. \textbf{Stage 1 (Idea Generation)} produces an idea $I$ that includes a brief method description and an experimental plan, and extends $I$ into a full research proposal $P$ that contains background, related work, method, experimental plan, and expected results. \textbf{Stage 2 (Experiment Execution)} validates $P$ by searching for and running executable code $C$ to yield verifiable outputs (e.g., logs and metrics) and to produce an execution report $W$.

\subsection{Overall Framework}
EvoScientist performs end-to-end scientific discovery for a \textbf{user goal} $G$ with three agents: a \textbf{researcher agent (RA)}, an \textbf{engineer agent (EA)}, and an \textbf{evolution manager agent (EMA)} (Figure~\ref{fig:framework}). For a given $G$, the RA first retrieves goal-relevant direction knowledge from an ideation memory $M_I$, generates an idea $I$, and extends it into a full proposal $P$. Conditioned on $P$, the EA retrieves reusable execution strategies from an experimentation memory $M_E$, searches for executable code $C$, runs experiments, and produces a verifiable execution report $W$ with outputs such as logs, metrics, and failure diagnoses. After the task is finished, the EMA summarizes the interaction histories to update $M_I$ (promising and failed directions) and $M_E$ (reusable execution strategies). For a new user goal, the RA and EA retrieve the updated memories before generating $I$ and $C$, enabling cross-task multi-agent evolution.

In the following subsections, we detail the researcher agent for idea tree search (Section~\ref{subsec:idea_tee}), the engineer agent for experiment tree search (Section~\ref{subsec:tree_search}), and the evolution manager agent (Section~\ref{subsec:evolution_manager}).

\subsection{Researcher Agent for Idea Tree Search}
\label{subsec:idea_tee}
To enable multi-agent evolution in idea generation, EvoScientist equips the researcher agent with a persistent ideation memory $M_I$ that records feasible directions and unsuccessful directions distilled from prior outcomes and failures.

\header{Ideation Memory Retrieval}
Given a user goal $G$, the researcher retrieves goal-relevant direction knowledge:
\begin{equation}
K_I = \text{Retrieve}_I(M_I, G),
\end{equation}
where $\text{Retrieve}_I(\cdot)$ is implemented by embedding-based retrieval with cosine distance similarity, and we select the top-$k_I$ most similar ideation memory items.

\header{Idea Tree Search}
Since the space of plausible ideas is large, the researcher agent performs a tree-structured propose--review--refine search grounded in literature review and retrieved memories. Concretely, each node in the search tree stores (i) an idea draft and (ii) its review feedback, and each expansion step uses the feedback to generate refined child ideas. The idea tree search generates a set of candidate ideas and their refinement signals:
\begin{equation}
\{(I_{1}, \text{rev}_{1}), \ldots, (I_{N_I}, \text{rev}_{N_I})\} = \text{IdeaTreeSearch}(G, L, K_I),
\end{equation}
where $I_{i}$ is the $i$-th candidate idea, $N_I$ is the maximum number of candidate ideas during tree search, and $\text{rev}_{i}$ stores the review feedback used in refinement, $L$ denotes the retrieved literature papers for $G$.

\header{Tournament Idea Selection}
EvoScientist uses an Elo-based tournament because it relies on pairwise comparisons and can produce a stable ranking under noisy judgments without requiring calibrated absolute scores.
The researcher ranks candidate ideas via an Elo-based tournament using idea quality (novelty, feasibility, relevance, and clarity):
\begin{equation}
\{r_{1}, \ldots, r_{N_I}\} = \text{EloRank}(I_{1:N_I}),
\end{equation}
where $r_{i}$ is the Elo rating score of idea $I_{i}$ after the tournament. We retain the top-$3$ ideas for direction summarization:
\begin{equation}
\mathcal{I}_{\text{top}} = \text{Top-}3(\{(I_{i}, r_{i})\}_{i=1}^{N_I}).
\end{equation}
Finally, the researcher extends the top-$1$ idea into a structured research proposal:
\begin{equation}
P = \text{Extend}(\text{Top-}1(\{(I_{i}, r_{i})\}_{i=1}^{N_I})).
\end{equation}
Here, the idea $I$ includes a method description and an experimental plan, while the proposal $P$ is a full version that contains background, related work, method, experimental plan, and expected results.

\subsection{Engineer Agent for Experiment Tree Search}
\label{subsec:tree_search}
To support multi-agent evolution in experiment execution, EvoScientist equips the engineer agent with a persistent experimentation memory $M_E$, which stores reusable data processing and model training strategies distilled from prior outcomes and failures.

\header{Experimentation Memory Retrieval}
Given a proposal $P$, the engineer retrieves reusable execution strategies and augments the base prompt:
\begin{equation}
K_E = \text{Retrieve}_E(M_E, P),
\end{equation}
where $\text{Retrieve}_E(\cdot)$ is implemented by embedding-based retrieval with cosine distance similarity, and we select the top-$k_E$ most similar experimentation memory items.

\header{Experiment Tree Search}
Because the space of implementations and execution environments is large, the engineer agent performs experiment tree search in four experiment stages (initial implementation, hyperparameter tuning, proposed method, and ablation). At each stage $s \in \{1,2,3,4\}$, it iteratively generates executable code, runs experiments, and records structured execution results; when execution fails, it diagnoses failures from logs and revises the code accordingly:
\begin{equation}
\{(C_{1}^s, E_{1}^s), \ldots, (C_{N_E^s}^{s}, E_{N_E^s}^{s})\} = \text{ExperimentTreeSearch}(P, K_E),
\end{equation}
where $C_{j}$ is $j$-th code at experiment stage $s$, $N_E^s$ is the maximum number of attempts for stage $s$, and $E_{j}^{s}$ is a structured execution record that includes run status, logs, and evaluation metrics. The best-performing code at each stage is selected as:
\begin{equation}
C_{best}^{s} = \text{argmax}_{j \in \{1,\ldots,N_E^s\}} \text{Top-}1(E_{j}^{s}),
\end{equation}
We define the execution history for each experiment stage $s$ as
\begin{equation}
H_E^s = \{(C_{j}^{s}, E_{j}^{s})\}_{j=1}^{N_E^s},
\end{equation}
and summarize execution outcomes into an execution report aligned with the proposal:
\begin{equation}
W = \text{SummarizeExecution}(P, \{H_E^s\}_{s=1}^{4}).
\end{equation}

\subsection{Evolution Manager Agent}
\label{subsec:evolution_manager}
The evolution manager agent (EMA) converts interaction histories into reusable strategies so that the system can learn from outcomes and failures and improve both idea generation and experiment execution across tasks. EvoScientist implements multi-agent evolution through three self-evolutions: idea direction evolution, idea validation evolution, and experiment strategy evolution.

\header{Idea Direction Evolution}
To accumulate reusable feasible directions, the EMA summarizes promising research directions from the top-ranked ideas $\mathcal{I}_{\text{top}}$:
\begin{equation}
F_I^{IDE} = \operatorname{IDE}(G,\mathcal{I}_{\text{top}}).
\end{equation}
where \(\operatorname{IDE}(\cdot)\) is implemented by prompting LLMs. Then, we update the ideation memory:
\begin{equation}
M_I \leftarrow \text{Update}_I(M_I, F_I^{IDE}).
\end{equation}

\header{Idea Validation Evolution}
To record unsuccessful directions from failures, the EMA analyzes the execution report $W$ for the selected proposal $P$. If the engineer cannot find any executable code within the pre-defined budget (rule-based), we treat the proposal as failed. Otherwise, when experiments are complete, the EMA compares the proposed method against baselines based on $W$ and uses an LLM-based analysis to judge whether the proposal fails (e.g., the proposed method performs worse than the baseline):
\begin{equation}
F_I^{IVE} = \operatorname{IVE}(P, W),
\end{equation}
where \(\operatorname{IVE}(\cdot)\) is implemented by prompting LLMs. Then, we use the idea validation analysis to update the ideation memory:
\begin{equation}
M_I \leftarrow \text{Update}_{I}(M_I, F_I^{IVE}).
\end{equation}

\header{Experiment Strategy Evolution}
To improve execution reliability, the EMA distills reusable execution strategies from the engineer's code search trajectories. The EMA summarizes reusable experiment strategies from both the best codes and  full trajectories in experiments:
\begin{equation}
F_E = \operatorname{ESE}(P,\{H_E^s\}_{s=1}^{4}),
\end{equation}
where \(\operatorname{ESE}(\cdot)\) is implemented by prompting LLMs. In this evolution, the EMA uses a prompt to jointly summarize (i) a \textbf{data processing strategy} and (ii) a \textbf{model training strategy}, and writes the resulting strategies into the experimentation memory $M_E$.
Finally, it updates the experimentation memory:
\begin{equation}
M_E \leftarrow \text{Update}_E(M_E, F_E).
\end{equation}

\vspace{-10pt}

\begin{table*}[htbp]
    \vspace{-1mm}
    \setlength{\tabcolsep}{4pt}
    \caption{Comparison of EvoScientist with baseline systems on scientific idea generation, evaluated by Gemini-3-flash. The scores marked with $\ast$ mean EvoScientist outperforms the baseline significantly with $p$-value < 0.05 (sign. test).}
    \label{tab:comparison_baselines}
    \centering \small
    \begin{tabular}{@{}lccccccccccccc@{}}
    \toprule
    & \multicolumn{3}{c}{\textbf{Novelty}} & \multicolumn{3}{c}{\textbf{Feasibility}} & \multicolumn{3}{c}{\textbf{Relevance}} & \multicolumn{3}{c}{\textbf{Clarity}} & \\
    \cmidrule(lr){2-4} \cmidrule(lr){5-7} \cmidrule(lr){8-10} \cmidrule(lr){11-13}
    \textbf{Method} & Win & Tie & Lose & Win & Tie & Lose & Win & Tie & Lose & Win & Tie & Lose & Avg. gap \\
    \midrule
    \multicolumn{14}{c}{\cellcolor{gray!15}\textit{Open-sourced Systems}} 
    \\
    EvoScientist vs Virtual Scientist & \textbf{96.67\rlap{$^{\ast}$}} & \phantom{0}3.33 & \phantom{0}0.00 & \textbf{93.33\rlap{$^{\ast}$}} & \phantom{0}6.67 & \phantom{0}0.00 & \textbf{90.00\rlap{$^{\ast}$}} & \phantom{0}6.67 & \phantom{0}3.33 & \textbf{96.67\rlap{$^{\ast}$}} & \phantom{0}3.33 & \phantom{0}0.00 & \textbf{+93.34} \\
    EvoScientist vs AI-Researcher & \textbf{96.67\rlap{$^{\ast}$}} & \phantom{0}3.33 & \phantom{0}0.00 & \textbf{90.00\rlap{$^{\ast}$}} & \phantom{0}0.00 & \phantom{0}10.00 & \textbf{86.67\rlap{$^{\ast}$}} & 10.00 & \phantom{0}3.33 & \textbf{93.34\rlap{$^{\ast}$}} & \phantom{0}3.33 & \phantom{0}3.33 & \textbf{+87.50} \\
    EvoScientist vs InternAgent & \textbf{73.33\rlap{$^{\ast}$}} & 16.67 & 10.00 & \textbf{93.33\rlap{$^{\ast}$}} & \phantom{0}0.00 & \phantom{0}6.67 & \textbf{86.67\rlap{$^{\ast}$}} & 13.33 & \phantom{0}0.00 & \textbf{96.67\rlap{$^{\ast}$}} & \phantom{0}3.33 & \phantom{0}0.00 & \textbf{+83.33} \\
    EvoScientist vs AI Scientist-v2 & \textbf{63.33\rlap{$^{\ast}$}} & 16.67 & 20.00 & \textbf{53.33\rlap{$^{\ast}$}} & \phantom{0}6.67 & 40.00 & \textbf{36.67\rlap{$^{\ast}$}} & 50.00 & 13.33 & \textbf{56.67\rlap{$^{\ast}$}} & 23.33 & 20.00 & \textbf{+29.17} \\
    \midrule
    \multicolumn{14}{c}{\cellcolor{gray!15}\textit{Commercial Systems}} 
    \\
    EvoScientist vs Hypogenic & \textbf{93.33\rlap{$^{\ast}$}} & \phantom{0}6.67 & \phantom{0}0.00 & \textbf{83.34\rlap{$^{\ast}$}} & \phantom{0}3.33 & 13.33 & \textbf{70.00\rlap{$^{\ast}$}} & 23.33 & \phantom{0}6.67 & \textbf{96.67\rlap{$^{\ast}$}} & \phantom{0}3.33 & \phantom{0}0.00 & \textbf{+80.83} \\
    EvoScientist vs Novix & \textbf{90.00\rlap{$^{\ast}$}} & \phantom{0}6.67 & \phantom{0}3.33 & \textbf{53.33\rlap{$^{\ast}$}} & 10.00 & 36.67 & \textbf{46.67\rlap{$^{\ast}$}} & 36.66 & 16.67 & \textbf{70.67\rlap{$^{\ast}$}} & 10.00 & 20.00 & \textbf{+46.00} \\
    EvoScientist vs K-Dense & \textbf{86.67\rlap{$^{\ast}$}} & \phantom{0}3.33 & 10.00 & \textbf{56.67\rlap{$^{\ast}$}} & 13.33 & 30.00 & \textbf{43.33\rlap{$^{\ast}$}} & 36.67 & 20.00 & \textbf{76.67\rlap{$^{\ast}$}} & 13.33 & 10.00 & \textbf{+54.50} \\
    \bottomrule
    \end{tabular}
    \vspace{-3mm}
  \end{table*}
  \begin{table*}[htbp]
  \vspace{-1mm}
    \caption{Comparison of EvoScientist with baseline systems on scientific idea generation, evaluated by human experts. The scores marked with $\ast$ mean EvoScientist outperforms the baseline significantly with $p$-value < 0.05 (sign. test).}
    \label{tab:comparison_baselines_human}
    \setlength{\tabcolsep}{4pt}
    \centering \small
    \begin{tabular}{@{}lccccccccccccc@{}}
    \toprule
    & \multicolumn{3}{c}{\textbf{Novelty}} & \multicolumn{3}{c}{\textbf{Feasibility}} & \multicolumn{3}{c}{\textbf{Relevance}} & \multicolumn{3}{c}{\textbf{Clarity}} & \\
    \cmidrule(lr){2-4} \cmidrule(lr){5-7} \cmidrule(lr){8-10} \cmidrule(lr){11-13}
    \textbf{Method} & Win & Tie & Lose & Win & Tie & Lose & Win & Tie & Lose & Win & Tie & Lose & Avg. gap \\
    \midrule
    \multicolumn{14}{c}{\cellcolor{gray!15}\textit{Open-sourced Systems}} 
    \\
    EvoScientist vs InternAgent & \textbf{66.67\rlap{$^{\ast}$}} & 23.33 & 10.00 & \textbf{96.67\rlap{$^{\ast}$}} & \phantom{0}3.33 & \phantom{0}0.00 & \textbf{90.00\rlap{$^{\ast}$}} & \phantom{0}0.00 & 10.00 & \textbf{93.33\rlap{$^{\ast}$}} & \phantom{0}6.67 & \phantom{0}0.00 & \textbf{+84.17} \\
    EvoScientist vs AI Scientist-v2 & \textbf{73.33\rlap{$^{\ast}$}} & 10.00 & 16.67 & \textbf{50.00\rlap{$^{\ast}$}} & 16.67 & 33.33 & \textbf{43.33\rlap{$^{\ast}$}} & 50.00 & 6.67 & \textbf{53.33\rlap{$^{\ast}$}} & 20.00 & 26.67 & \textbf{+34.16} \\
    \midrule
    \multicolumn{14}{c}{\cellcolor{gray!15}\textit{Commercial Systems}} 
    \\
    EvoScientist vs Novix & \textbf{93.33\rlap{$^{\ast}$}} & \phantom{0}0.00 & \phantom{0}6.67 & \textbf{56.67\rlap{$^{\ast}$}} & 6.66 & 36.67 & \textbf{36.67\rlap{$^{\ast}$}} & 60.00 & \phantom{0}3.33 & \textbf{73.33\rlap{$^{\ast}$}} & 10.00 & 16.67 & \textbf{+49.17} \\
    EvoScientist vs K-Dense & \textbf{96.67\rlap{$^{\ast}$}} & \phantom{0}3.33 & \phantom{0}0.00 & \textbf{53.33\rlap{$^{\ast}$}} & 26.67 & 20.00 & \textbf{40.00\rlap{$^{\ast}$}} & 43.33 & 16.67 & \textbf{53.34\rlap{$^{\ast}$}} & 43.33 & \phantom{0}3.33 & \textbf{+50.84} \\
    \bottomrule
    \end{tabular}
    \vspace{-3mm}
  \end{table*}
\section{Experimental Setup}
\label{sec:experimental_setup}

\subsection{Research Questions}
To empirically validate the EvoScientist framework, we design experiments that evaluate its performance across different stages of the scientific discovery pipeline, following the core claims introduced in Section~\ref{sec:introduction}. We address the following research questions:

\begin{enumerate}[label=\textbf{(RQ\arabic*)}, leftmargin=*,topsep=2pt, itemsep=2pt]
    \item How well does EvoScientist generate high-quality scientific ideas in terms of novelty, feasibility, relevance, and clarity?
    \item How reliable is EvoScientist in generating and executing experimental code, as measured by execution success rates?
    \item How well does EvoScientist perform in end-to-end scientific discovery tasks, from idea generation to producing publication-quality research papers?
    \item To what extent does the proposed multi-agent evolution mechanism contribute to improvement in idea quality?
\end{enumerate}

\subsection{Datasets}

Since we focus on end-to-end scientific discovery, no publicly available dataset covers the complete pipeline. To comprehensively evaluate EvoScientist across different stages of the scientific discovery pipeline, we construct a multi-level evaluation set covering three core tasks: idea generation, code implementation, and end-to-end scientific discovery.

\begin{itemize}[leftmargin=*,topsep=2pt, itemsep=2pt,label=$\circ$]
    \item \textbf{Idea Generation:} We curate 30 research queries solicited from experienced AI researchers, spanning diverse contemporary topics in artificial intelligence. Following recent benchmarks~\citep{liu2025researchbench,su2025many}, this set is used to evaluate idea quality along four dimensions: novelty, feasibility, relevance, and clarity.
    \item \textbf{Code Generation:} For each research query, the corresponding research proposal generated in the idea generation stage serves as input, evaluating the system's ability to implement and execute experiments that operationalize the proposed ideas.
    \item \textbf{End-to-End Scientific Discovery:} We select 6 research ideas and develop them into complete research manuscripts, which are submitted to the International Conference on AI Scientists (ICAIS~2025~\citep{icais2025}) for peer review. Following prior work~\citep{yamada2025ai}, manuscript writing uses the same paper-writing module as AI Scientist-v2, as our evolution mechanism focuses on ideation and experiment execution rather than scientific writing.
\end{itemize}
Dataset details (queries, tasks, and selection) are in Appendix~\ref{appendix:data}.

\subsection{Baselines}
To comprehensively evaluate EvoScientist, we compare it with representative open-sourced systems---Virtual Scientist~\citep{su2025many}, AI-Researcher~\citep{tang2025ai}, InternAgent~\citep{team2025novelseek}, and AI Scientist-v2~\citep{yamada2025ai}---and commercial systems---Hypogenic~\citep{hypogenic2026}, Novix~\citep{novix2026}, and K-Dense~\citep{kdense2026}. Detailed baseline descriptions are provided in Appendix~\ref{appendix:base}.

\subsection{Evaluation Metrics}
We evaluate EvoScientist across three core tasks using a combination of automated LLM-based evaluations and expert human judgments, following established practices in recent work on autonomous scientific discovery~\citep{lu2024ai,yamada2025ai}.
\begin{itemize}[leftmargin=*,topsep=2pt, itemsep=2pt,label=$\circ$]
    \item \textbf{Idea Generation}: Idea quality is assessed via pairwise comparisons conducted by both an LLM judge and expert human evaluators~\citep{liu2025researchbench,su2025many}. For LLM evaluation, each comparison is evaluated in swapped positions to reduce positional bias. Judges score ideas on a $1$--$10$ scale along four dimensions—\textit{Novelty}, \textit{Feasibility}, \textit{Relevance}, and \textit{Clarity}—with outcomes aggregated into \emph{Win}, \emph{Tie}, or \emph{Lose}. For human evaluation, we recruit three PhD-level annotators in relevant AI areas. Annotators can use the internet to verify literature-related claims when necessary.

    \item \textbf{Code Generation}: Code generation performance is measured by the execution success rate, defined as the proportion of trials in which generated code executes successfully in a sandboxed environment and produces valid outputs~\citep{lu2024ai,yamada2025ai}.

    \item \textbf{End-to-End Scientific Discovery}: End-to-end performance is evaluated through academic peer review of the six manuscripts submitted to ICAIS~2025~\citep{icais2025}, incorporating assessments from both an automated AI reviewer and official human reviewers.
\end{itemize}
More evaluation details and scoring rubrics are in Appendix~\ref{appendix:eval}.

\subsection{Implementation Details}
Our implementation leverages a set of state-of-the-art language models and external tools tailored to different stages of the scientific discovery pipeline. For the initial literature review phase, we utilize the Semantic Scholar API to retrieve relevant papers. Scientific idea generation is performed using Gemini-2.5-Pro. For the code generation task, we employ Claude-4.5-Haiku. End-to-end manuscript authoring is also handled by Gemini 2.5 Pro to ensure coherence and high-quality scientific writing. 
For memory indexing and retrieval, we employ the `mxbai-embed-large'~\citep{emb2024mxbai} embedding model via Ollama~\citep{ollama2026}.
We set the ideation retrieval top-$k_I$ to $2$ and use a maximum of $N_I=21$ candidate ideas during idea tree search, with $3$ parallel workers. For experiment execution, we set the experimentation retrieval top-$k_E$ to $1$ and use $4$ parallel workers. The maximum numbers of attempts at stages $s\in\{1,2,3,4\}$ are $N_E^1=20$, $N_E^2=12$, $N_E^3=12$, and $N_E^4=18$. 

All experiments are conducted under consistent task settings and evaluation protocols across EvoScientist and baseline systems, with additional baseline details provided in Appendix~\ref{appendix:base} and additional implementation details provided in Appendix~\ref{appendix:training}.

\section{Experimental Results}
\label{sec:results}

\subsection{Idea Generation Performance (RQ1)}
\header{Automatic evaluation}
Table~\ref{tab:comparison_baselines} presents pairwise comparison results for scientific idea generation, evaluated by an advanced LLM judge (Gemini-3-flash). EvoScientist is compared against both open-sourced and commercial AI scientist systems across four dimensions: \textbf{Novelty}, \textbf{Feasibility}, \textbf{Relevance}, and \textbf{Clarity}. These results provide evidence for the effectiveness of EvoScientist's idea generation capability under automatic evaluation. We highlight three main observations:
\begin{itemize}[leftmargin=*,topsep=2pt, itemsep=2pt]
            \item \textbf{EvoScientist outperforms both open-sourced and commercial baselines on the overall average gap.}
        Table~\ref{tab:comparison_baselines} shows that the Avg. gap is positive across all comparisons against both open-sourced systems and commercial systems, indicating an overall advantage aggregated over novelty, feasibility, relevance, and clarity. In particular, the Avg. gap ranges from +29.17 to +93.34 against open-sourced baselines and from +46.00 to +80.83 against commercial baselines.

            \item \textbf{Compared to open-sourced and commercial baselines, EvoScientist achieves stronger performance on novelty and feasibility.}
        As shown in Table~\ref{tab:comparison_baselines}, EvoScientist wins more frequently than open-sourced and commercial baselines on both novelty and feasibility. This trend aligns with the memory-driven multi-agent evolution design: the evolution manager distills outcomes and failures into ideation memory, which the researcher agent retrieves and incorporates into subsequent prompts, thereby improving the feasibility and originality of generated ideas over time.

        \item \textbf{EvoScientist improves relevance and clarity against both open-sourced and commercial baselines via idea tree search and tournament selection.}
        Table~\ref{tab:comparison_baselines} shows that EvoScientist achieves strong performance on relevance, while the largest performance gaps appear in the clarity dimension across a wide range of baselines. This pattern is aligned with the propose--review--refine idea tree search, which generates candidate ideas together with explicit critique signals, and the subsequent Elo-based tournament ranks candidates using novelty, feasibility, relevance, and clarity.
\end{itemize}

\header{Human evaluation}
\label{ssec:multi_judgment_prediction_results}
Human evaluation provides a more reliable assessment of scientific idea quality. Table~\ref{tab:comparison_baselines_human} reports pairwise results from expert human judges. To ensure a focused yet challenging comparison, we selected baselines that demonstrated strong performance during automatic evaluation.
\begin{itemize}[leftmargin=*,topsep=2pt, itemsep=2pt]
        \item \textbf{EvoScientist consistently outperforms strong baselines in terms of novelty and feasibility under human evaluation.} 
        As shown in Table~\ref{tab:comparison_baselines_human}, expert judges prefer EvoScientist more often than strong open-source and commercial baselines on both novelty and feasibility, with win rates consistently exceeding lose rates across comparisons. Averaged across four representative comparisons (InternAgent, AI Scientist-v2, Novix, and K-Dense), EvoScientist achieves a Novelty win rate of 82.50\% and a Feasibility win rate of 64.17\%.

        \item \textbf{EvoScientist remains competitive on relevance and achieves stronger advantages on clarity under human evaluation.}
        As shown in Table~\ref{tab:comparison_baselines_human}, relevance exhibits a higher rate of \emph{Ties}, particularly against commercial baselines, suggesting that topical alignment can be subtle and evaluator-dependent. Nevertheless, EvoScientist’s win rates on relevance consistently exceed its lose rates, while its clarity wins are more pronounced. This pattern is consistent with the idea tree search and Elo-based tournament selection, which favor ideas that are more explicit, making differences in presentation quality easier to assess even when relevance judgments converge.
\end{itemize}

\begin{figure}[t]
\vspace{-2pt}
  \centering
\includegraphics[width=0.8\linewidth]{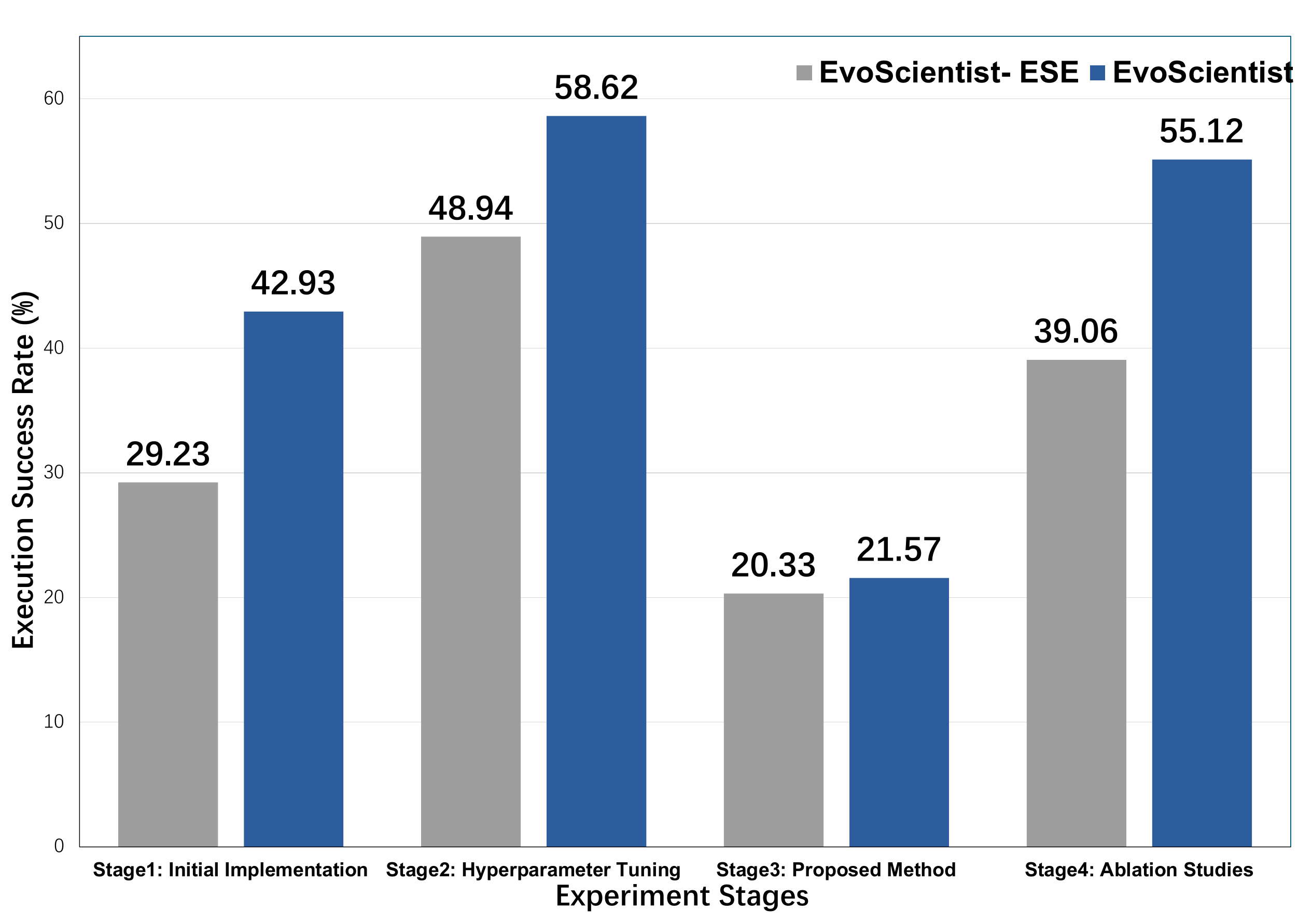}
\vspace*{-2mm}
\caption{Mean execution success rate across four experiment stages, before and after experiment strategy evolution (ESE).} 
\vspace*{-5mm}

\label{fig:codegen}
\end{figure}

\begin{table*}[htbp]
  \vspace{-1mm}
  \caption{Ablation study on scientific idea generation, evaluated by Gemini-3-flash.} 
      \label{tab:ablation_study}
      \setlength{\tabcolsep}{4pt}
      \centering \small
      \begin{tabular}{@{}lccccccccccccc@{}}
      \toprule
      & \multicolumn{3}{c}{\textbf{Novelty}} & \multicolumn{3}{c}{\textbf{Feasibility}} & \multicolumn{3}{c}{\textbf{Relevance}} & \multicolumn{3}{c}{\textbf{Clarity}} & \\
      \cmidrule(lr){2-4} \cmidrule(lr){5-7} \cmidrule(lr){8-10} \cmidrule(lr){11-13}
      \textbf{Method} & Win & Tie & Lose & Win & Tie & Lose & Win & Tie & Lose & Win & Tie & Lose & Avg. gap \\
      \midrule
      -IDE vs EvoScientist & 16.67 & 16.67 & 66.67 & 20.00 & 30.00 & 50.00 & 23.33 & 50.00 & 26.67 & 23.33 & 46.67 & 30.00 & -22.50 \\
      -IVE vs EvoScientist & 30.00 & 26.67 & 43.33 & 10.00 & 26.67 & 63.33 & 30.00 & 46.67 & 23.33 & 16.67 & 46.67 & 36.67 & -20.00 \\
      -all vs EvoScientist & 10.00 & 10.00 & 80.00 & \phantom{0}3.33 & 13.33 & 83.33 & 16.67 & 46.67 & 36.67 & 20.00 & 46.67 & 33.33 & -45.83 \\
      \bottomrule
      \end{tabular}
      \vspace*{-3mm}
\end{table*}

\subsection{Code Generation Performance (RQ2)}
We evaluate code generation of EvoScientist using the execution success rate, and report results before and after experiment strategy evolution (ESE) across four experiment stages in Figure~\ref{fig:codegen}. Based on the results, we highlight two main observations:
\begin{itemize}[leftmargin=*,topsep=2pt, itemsep=2pt]
    \item \textbf{EvoScientist improves the mean execution success rate across four experiment stages after experiment strategy evolution.}
    Averaged across all stages, EvoScientist’s execution success rate increases from 34.39 before evolution to 44.56 after evolution. This gain is consistent with experiment strategy evolution: the evolution manager distills outcomes and failures from code-generation trajectories into experimentation memory, which the engineer agent retrieves and incorporates into subsequent prompts to produce more reliable implementations.

    \item \textbf{EvoScientist achieves measurable progress on the proposed method stage (stage 3) after evolution, while this stage remains challenging.}
    In stage 3, EvoScientist’s mean execution success rate increases from 20.33 to 21.57 after evolution. Although the improvement is modest, it indicates that experiment strategy evolution can still accumulate and reuse execution lessons in a difficult setting. The remaining low success rate also highlights clear headroom for improvement, suggesting that richer interaction histories and more fine-grained execution feedback may further strengthen EvoScientist’s performance on complex proposed-method implementations.
\end{itemize}

\subsection{End-to-end Scientific Discovery Performance (RQ3)}
To evaluate end-to-end scientific discovery, we deployed EvoScientist to autonomously generate and write six complete research papers, which are subsequently submitted to the ICAIS 2025~\citep{icais2025}. 
The AI Scientist Track received 82 submissions and accepted 26 papers, corresponding to an acceptance rate of 31.71\%. As shown in Table~\ref{tab:icais_results} (Appendix~\ref{appendix:case_study_detail}), all six papers generated by EvoScientist were accepted. Among them, one paper received the \textbf{\textit{Best Paper Award}}, and another received the \textbf{\textit{AI Reviewer's Appraisal Award}}.

Beyond acceptance outcomes, we further analyze peer review feedback to characterize EvoScientist’s end-to-end behavior.  We synthesize meta-reviews from all six accepted paper, revealing a consistent profile of strengths and limitations that align with the proposed memory-driven multi-agent evolution framework.
\begin{itemize}[leftmargin=*,topsep=2pt, itemsep=2pt]
    \item \textbf{EvoScientist demonstrates strength in methodological novelty.}
    Reviewers consistently highlighted the novelty, relevance, and clarity of the research problems addressed by EvoScientist-generated submissions. This is consistent with the researcher agent's use of ideation memory, where reusable ideation strategies distilled from prior outcomes and failures are retrieved and incorporated into subsequent proposal generation.

    \item \textbf{EvoScientist delivers robust experimental validation and execution.}
    Four of the six papers received explicit praise for their "comprehensive and sound experimental design" or "solid empirical evidence." This reflects the engineer agent's ability to implement and execute experiments with support from experimentation memory, which stores reusable execution lessons, such as debugging patterns and validated implementations, and systematic reuse across tasks.

    \item \textbf{EvoScientist can further benefit from stronger theoretical analysis and formal grounding.}
    A recurring critique concerned the lack of deeper theoretical formalization beyond empirical results. EvoScientist prioritizes generating testable proposals and producing experimental evidence through outcome-driven evolution. As a result, the system does not consistently abstract empirical findings into formal theoretical frameworks, defining a clear handover point: EvoScientist delivers the empirical findings (the ``what''), while deeper theoretical interpretation (the ``why'') remains a direction for human researchers.
\end{itemize}
More case studies and feedback are provided in Appendix~\ref{appendix:case_study_detail}.
  
\subsection{Ablation Studies (RQ4)}
\label{ssec:ablation_study}
In Table~\ref{tab:ablation_study}, we compare EvoScientist with several ablative variants. The variants are as follows:
\begin{enumerate*}[label=(\roman*)]
\item \textbf{-IDE}: we remove idea direction evolution. 
\item \textbf{-IVE}: we remove idea validation evolution.
\item \textbf{-all}: we remove all idea evolution. 
\end{enumerate*}
Our findings are as follows:
\begin{itemize}[leftmargin=*,topsep=2pt, itemsep=2pt]
  \item \textbf{Removing idea direction evolution reduces both novelty and feasibility.}
  When removing idea direction evolution (\textbf{-IDE}), the ablative variant loses to EvoScientist more frequently on both Novelty (Lose: 66.67\%) and Feasibility (Lose: 50.00\%). This suggests that evolving and reusing direction-level insights plays a crucial role in guiding the researcher agent toward ideas that are not only more original but also more practically grounded.

  \item \textbf{Removing idea validation evolution disproportionately harms feasibility.}
  Removing idea validation evolution (\textbf{-IVE}) leads to a larger degradation in Feasibility, with the variant losing to EvoScientist on Feasibility in 63.33\% of comparisons. This result suggests that validation-driven evolution is essential for filtering out experimentally infeasible directions.

  \item \textbf{Removing all idea evolution causes substantial drops in novelty and feasibility, but smaller changes in relevance and clarity.}
  When removing all idea evolution (\textbf{-all}), performance degrades on Novelty (Lose: 80.00\%) and Feasibility (Lose: 83.33\%). In contrast, the differences are less pronounced on Relevance and Clarity, where a large fraction of comparisons result in ties (both 46.67\%). This pattern indicates that the primary benefits of idea evolution arise from improving originality and feasibility, rather than surface-level relevance of linguistic clarity.
\end{itemize}



\section{Conclusions}
\label{sec:conclusion}
In this paper, we introduced EvoScientist, a multi-agent evolving framework that addresses a core limitation of existing end-to-end AI scientist systems: most rely on static, hand-designed pipelines and do not adapt their idea- or code-generation strategies from accumulated interaction histories. EvoScientist enables continuous improvement through persistent memory and self-evolution, coordinating three specialized agents: a Researcher Agent (\textit{RA}) for scientific idea generation, an Engineer Agent (\textit{EA}) for experiment implementation and execution, and an Evolution Manager Agent (\textit{EMA}) that distills insights from prior agent interactions into reusable knowledge.

EvoScientist maintains two persistent memory modules: (i) an ideation memory that summarizes feasible research directions from top-ranked ideas while recording previously unsuccessful directions identified during idea validation; and (ii) an experimentation memory that captures effective data processing and model training strategies derived from code search trajectories and best-performing implementations. By retrieving relevant prior strategies from these memories, the \textit{RA} and \textit{EA} improve idea quality and increase code execution success rates over time. Experiments show that EvoScientist outperforms \textbf{7} open-source and commercial state-of-the-art systems in scientific idea generation, achieving higher performance in novelty, feasibility, relevance, and clarity under both automatic and human evaluation, and it substantially improves code execution success rates through multi-agent evolution.

\vspace{-10pt}
\section{Limitations and Ethical Considerations}

EvoScientist is designed to support human-led scientific discovery. We summarize key limitations and ethical considerations.

\header{Limitations}
Our evaluation focuses on computational research tasks~\cite{shi2025deep,lyu2025deepshop,wang2025cooperative,hao2025token,hao2025uni,lyu2024knowtuning} where hypotheses can be tested via simulation and code execution. Generalization to domains that require physical experimentation (e.g., materials science and drug discovery) remains open, and will require integration with laboratory workflows and real-world feedback.

\header{Ethical Considerations}
EvoScientist should be used as a decision support system~\cite{DBLP:journals/ipm/LyuWRRCLLLS22,DBLP:conf/emnlp/LyuH0ZGRCWR23,zhang2024towards} rather than a replacement for expert judgment. All outputs should be verified by human experts before dissemination, and deployments should include safeguards against dual-use and misuse~\cite{lyumacpo}. Since the system learns from existing literature, it may reproduce biases in data and writing~\cite{lyu2024cognitive,lyu2025self,DBLP:conf/aaai/LyuLYRRZYR23}, and should be monitored and audited accordingly.

\section{GenAI Disclosure}
In this work, GenAI is used exclusively as a general-purpose tool for language refinement and manuscript polishing. 
LLMs did not contribute to the conceptualization, experimental design, data analysis, or interpretation of the results in this work. 
All scientific content, findings, and conclusions presented in this paper are the sole responsibility of the authors. 
No text generated by LLMs affects the originality or intellectual contribution of the work. 


\bibliographystyle{ACM-Reference-Format}
\bibliography{references}

\appendix
\section*{Appendix}
\section{Details of Datasets}
\label{appendix:data}
We curate a set of $30$ research queries solicited from experienced AI researchers. Before evaluation, we rewrite each query into a unified template for consistency. During evaluation, each query string is used verbatim as the user goal $G$.

\begin{figure*}[htbp]
  \centering
\includegraphics[width=0.9\textwidth]{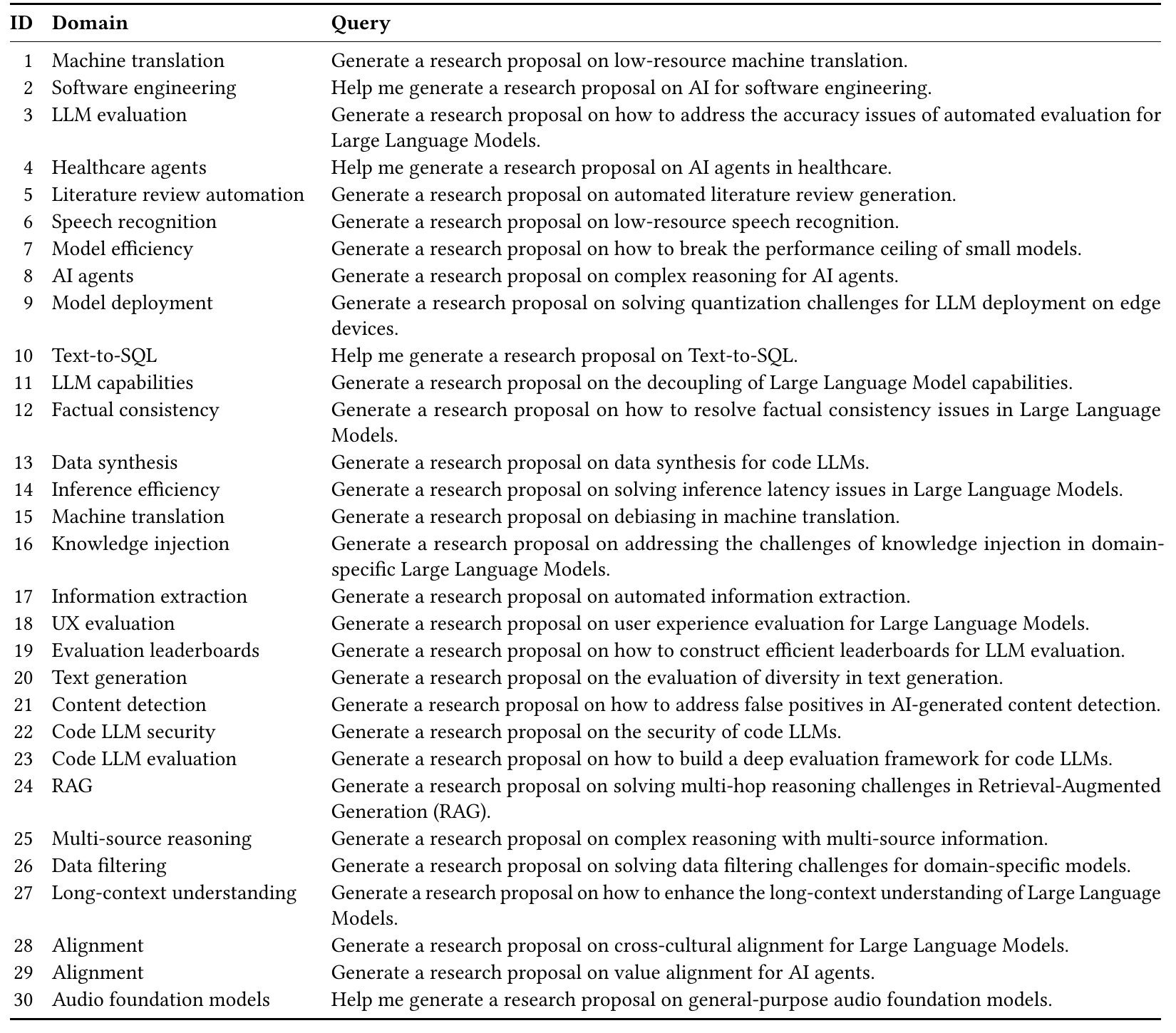}
\vspace*{-1mm}
\caption{Research queries used for evaluation.} 
\vspace*{-3mm}
\label{fig:eval_queries}
\end{figure*}

The queries cover multiple areas, including machine translation, speech recognition, software engineering, healthcare agents, literature review automation, text-to-SQL, information extraction, retrieval-augmented generation, multimodal architectures, model efficiency and deployment, data synthesis, safety, and alignment. We list the $30$ queries in Figure~\ref{fig:eval_queries}.

\section{Details of Baselines}
\label{appendix:base}

We compare EvoScientist with both open-sourced and commercial systems that represent strong baselines for autonomous scientific discovery. Where applicable, baseline methods are configured following the settings reported in their original papers or official documentation.

\textbf{\textit{Open-sourced systems.}}
\begin{itemize}[leftmargin=*,topsep=2pt, itemsep=2pt,label=$\circ$]
    \item \textbf{Virtual Scientist}~\citep{su2025many} is an \ac{LLM}-based multi-agent system that simulates collaborative scientific ideation through proposal, critique, and refinement, representing a strong baseline for idea generation.
    \item \textbf{AI-Researcher}~\citep{tang2025ai} is a fully autonomous research system that orchestrates the complete research pipeline, from literature review and hypothesis generation to experiment implementation and manuscript preparation, with minimal human intervention.
    \item \textbf{InternAgent}~\citep{team2025novelseek} is a closed-loop multi-agent framework for autonomous scientific research, emphasizing scalability, interactivity, and human-in-the-loop extensibility.
    \item \textbf{AI Scientist-v2}~\citep{yamada2025ai} is an end-to-end agentic system that employs a progressive agentic tree-search methodology to autono-mously generate hypotheses, design experiments, analyze data, and produce scientific manuscripts.
\end{itemize}

\textbf{\textit{Commercial systems.}}
\begin{itemize}[leftmargin=*,topsep=2pt, itemsep=2pt,label=$\circ$]
    \item \textbf{Hypogenic}~\citep{hypogenic2026} is a community-driven AI research acceleration platform that utilizes AI agents to assist in interdisciplinary scientific exploration, featuring a weekly competition where winning ideas are implemented by AI research agents.
    \item \textbf{Novix}~\citep{novix2026} is an AI Agentic Co-scientist designed to accelerate the full research lifecycle, including idea generation, literature review, experimental design, and data analysis.
    \item \textbf{K-Dense}~\citep{kdense2026} is an agentic AI platform positioned as an intelligent task executor, supporting end-to-end automation from data processing to decision-making across multiple domains.
\end{itemize}

\section{Details of Evaluation}
\label{appendix:eval}
\subsection{LLM Evaluation for Idea Generation}
\label{appendix:gpt4}
\begin{figure*}[htbp]
  \centering
\includegraphics[width=0.8\textwidth]{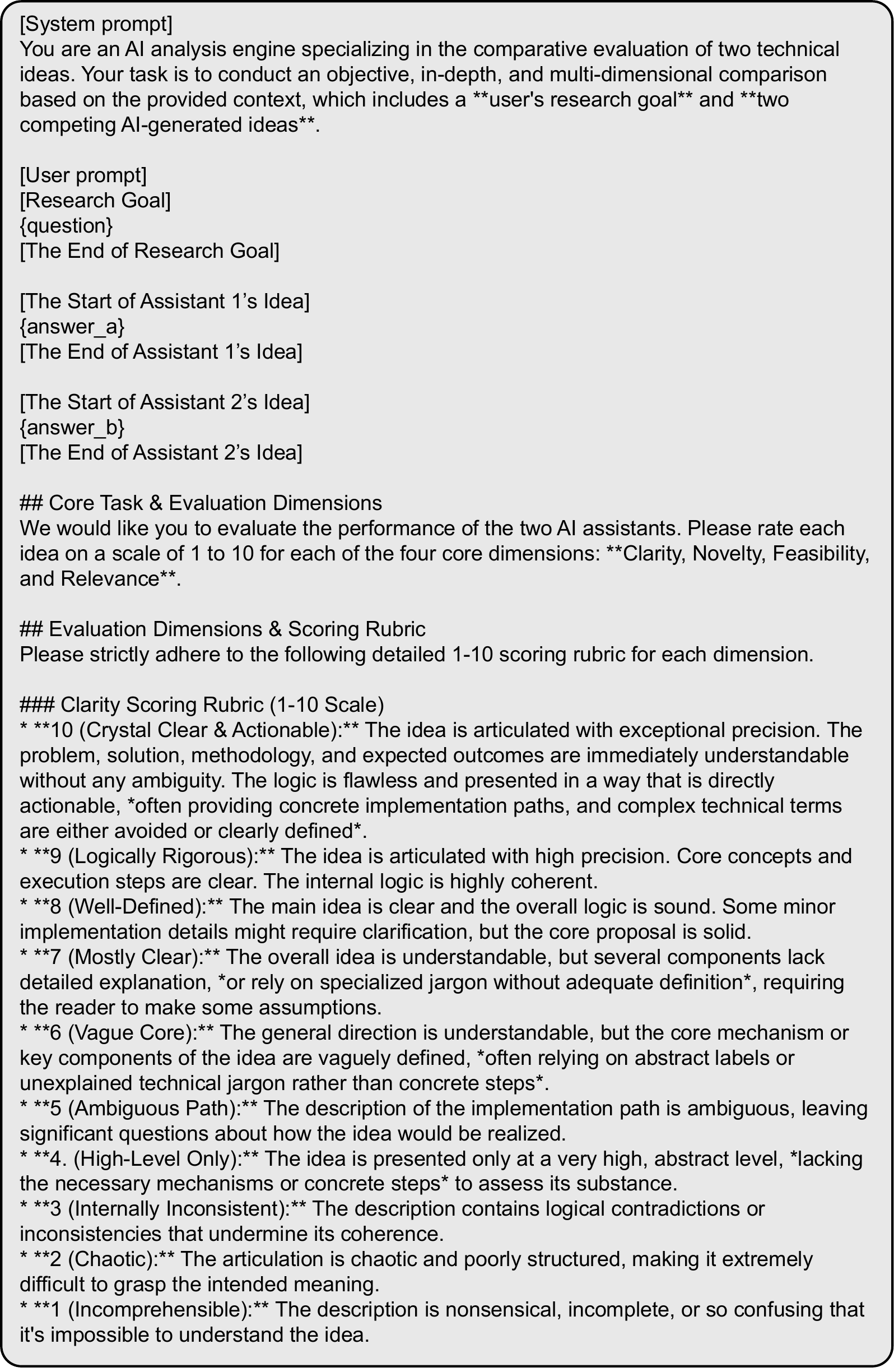}
\vspace*{-1mm}
\caption{Part 1. Prompt for LLM-based idea generation evaluation.} 
\vspace*{-3mm}
\label{fig:llm_eval1}
\end{figure*}

\begin{figure*}[htbp]
  \centering
\includegraphics[width=0.8\textwidth]{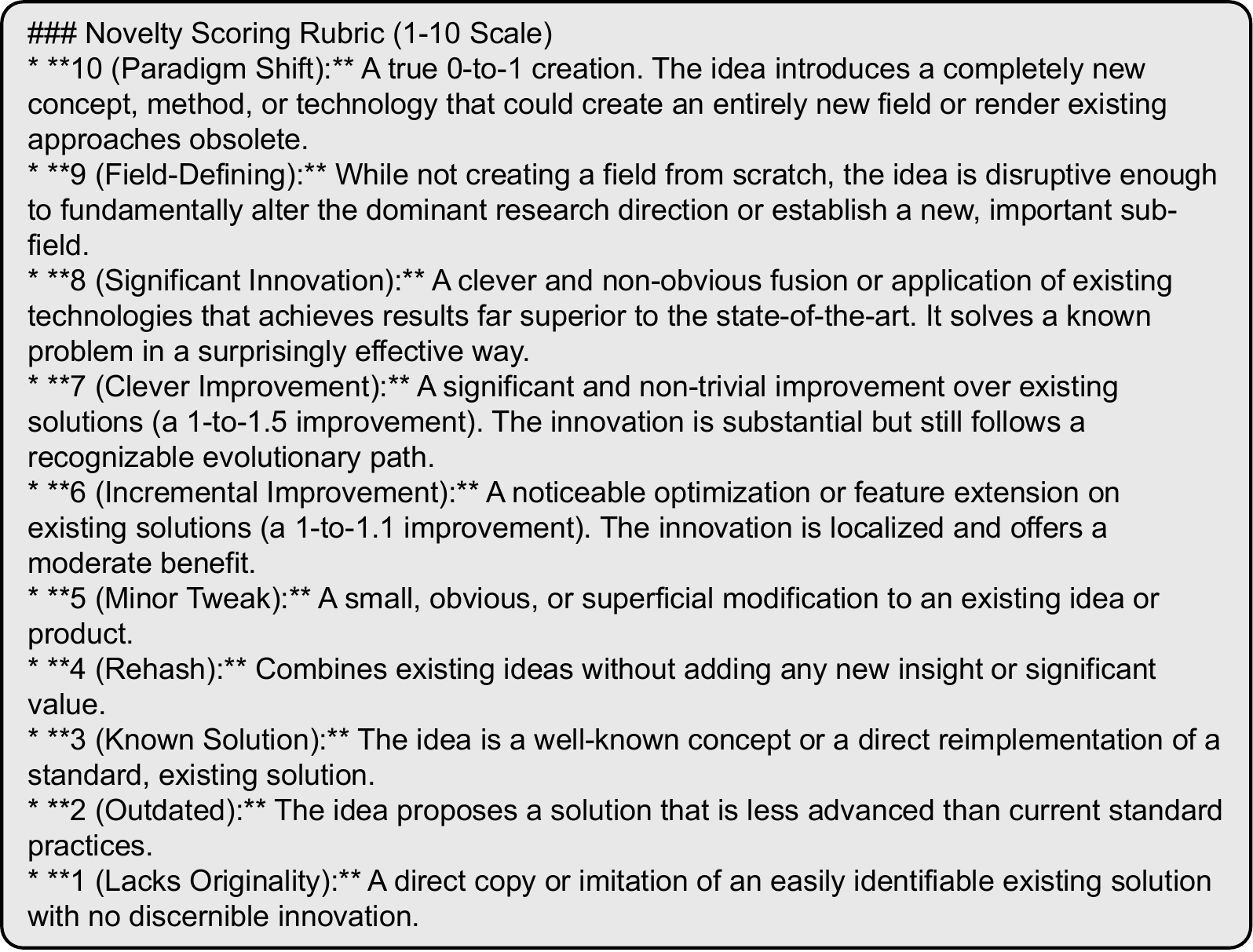}
\caption{Part 2. Prompt for LLM-based idea generation evaluation.} 
\vspace*{-3mm}
\label{fig:llm_eval2}
\end{figure*}

\begin{figure*}[htbp]
  \centering
\includegraphics[width=0.8\textwidth]{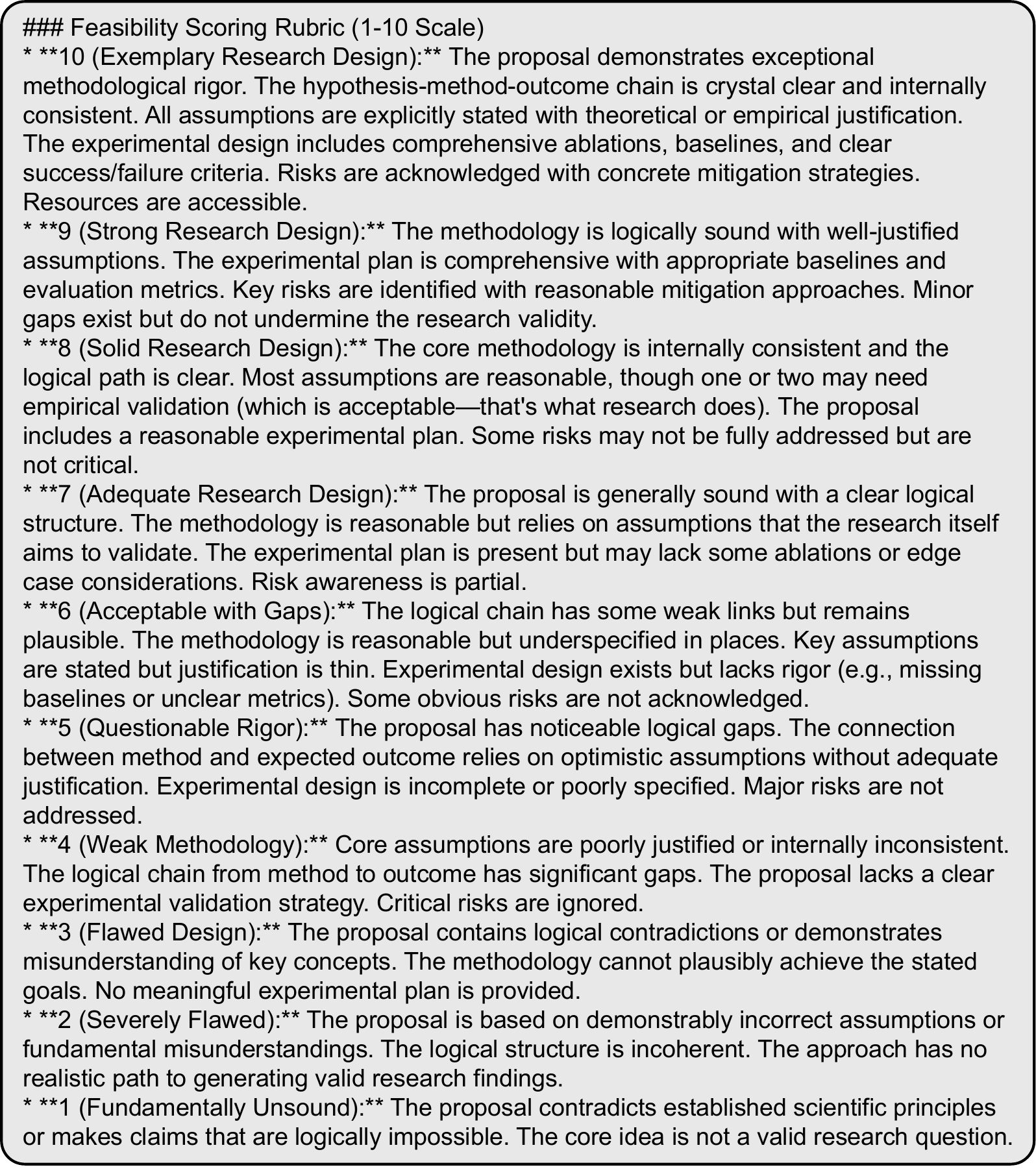}
\vspace*{-1mm}
\caption{Part 3. Prompt for LLM-based idea generation evaluation.} 
\vspace*{-3mm}
\label{fig:llm_eval3}
\end{figure*}

\begin{figure*}[htbp]
  \centering
\includegraphics[width=0.8\textwidth]{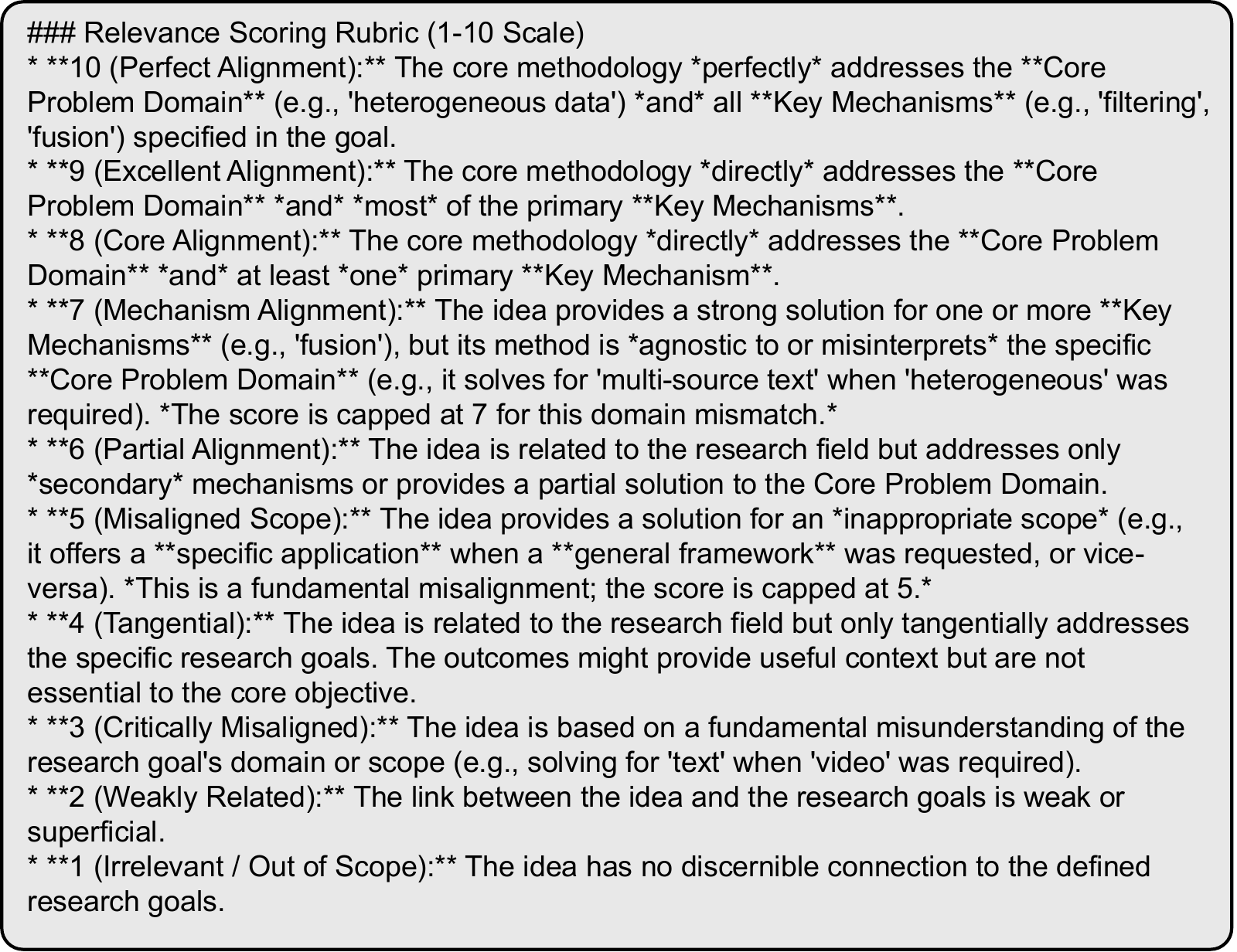}
\vspace*{-1mm}
\caption{Part 4. Prompt for LLM-based idea generation evaluation.} 
\vspace*{-3mm}
\label{fig:llm_eval4}
\end{figure*}

\begin{figure*}[htbp]
  \centering
\includegraphics[width=0.6\textwidth]{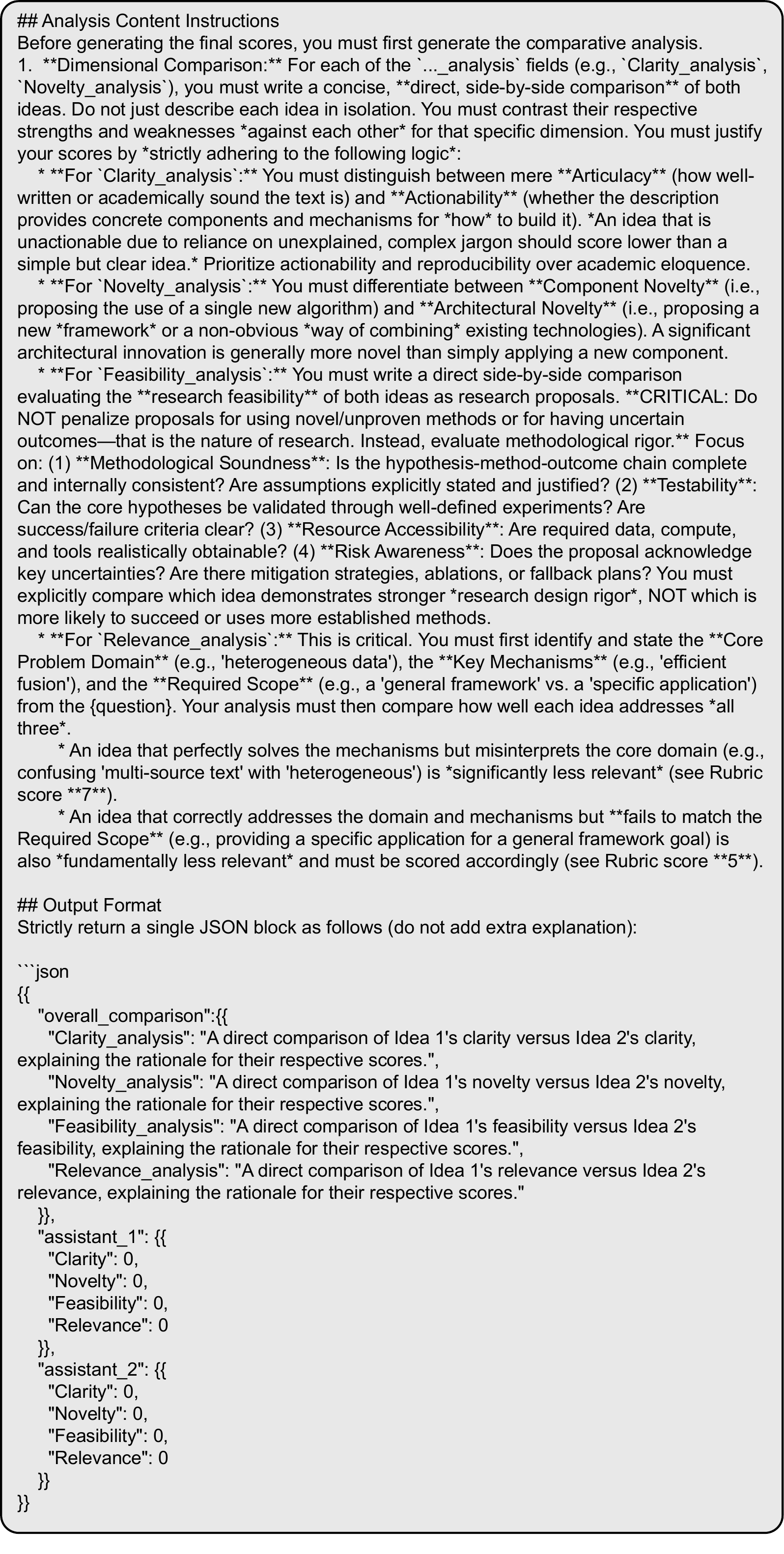}
\vspace*{-1mm}
\caption{Part 5. Prompt for LLM-based idea generation evaluation.} 
\vspace*{-3mm}
\label{fig:llm_eval5}
\end{figure*}

This subsection reports the prompt used to obtain pairwise LLM judgments for evaluating idea-generation outputs with \textit{gemini-3-flash}. Figures~\ref{fig:llm_eval1}--\ref{fig:llm_eval5} present the prompt template adapted from \citet{zheng2024judging}, which assesses four dimensions: novelty, feasibility, relevance, and clarity. To mitigate positional bias~\cite{DBLP:conf/emnlp/KoLKKK20,DBLP:journals/corr/abs-2305-17926}, we evaluate each answer pair twice by swapping their order in two independent runs.

\begin{figure*}[htbp]
  \centering
\includegraphics[width=0.9\textwidth]{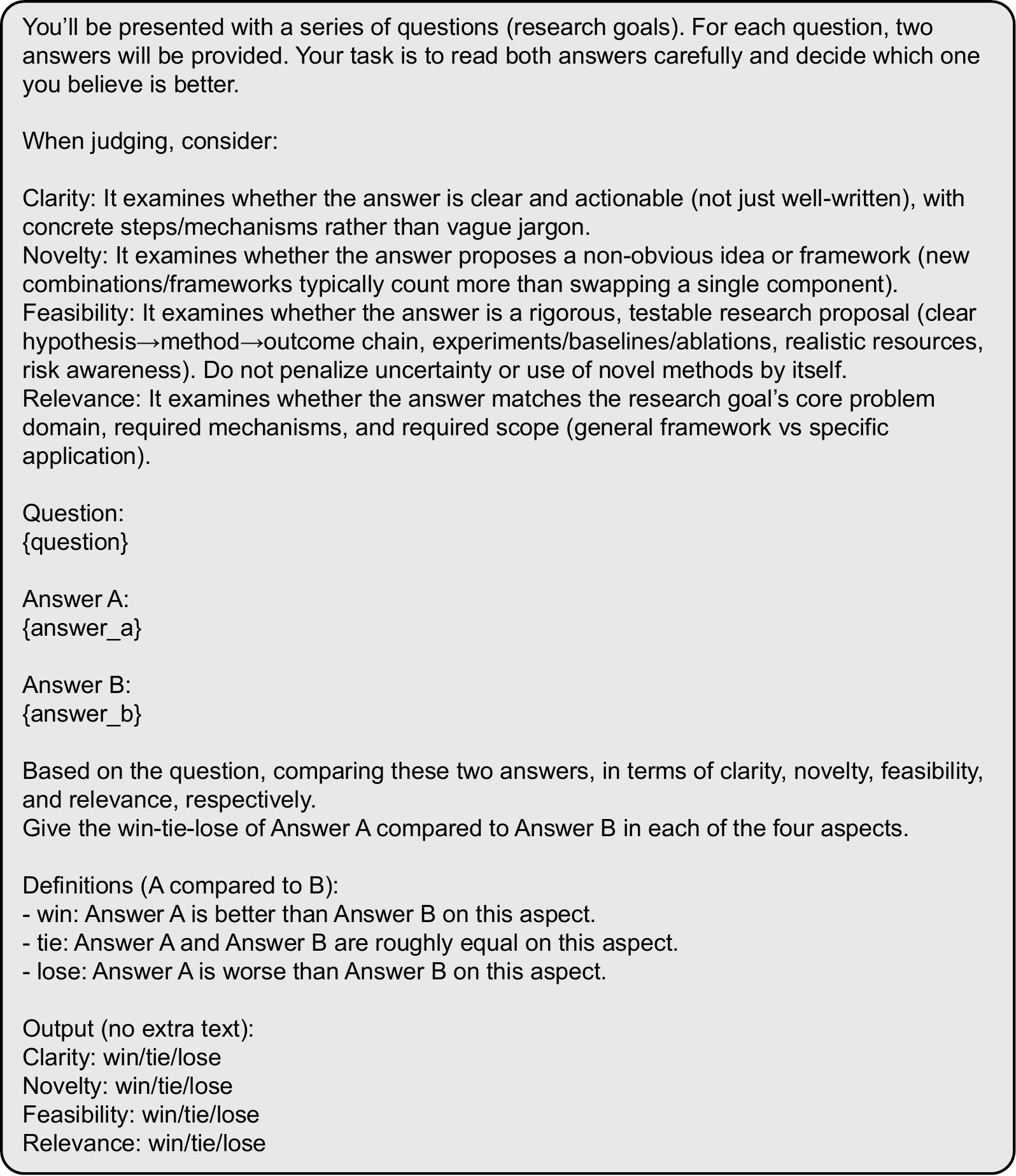}
\vspace*{-1mm}
\caption{Instructions for human evaluation of idea generation.} 
\vspace*{-3mm}
\label{fig:human_eval}
\end{figure*}

\subsection{Human Evaluation for Idea Generation}
\label{appendix:human}
For human evaluation, we recruited 3 annotators with PhD degrees in relevant AI domains to do human evaluation. Annotators were allowed to use the internet to verify literature-related claims when needed. The full instructions are shown in Figure~\ref{fig:human_eval}.

\subsection{Agreement between LLM Evaluation and Human Evaluation for Idea Generation}
\label{appendix:human}
To validate the reliability of our LLM-based evaluation, we conducted a human evaluation study on a subset of 120 idea pairs (30 pairs × 4 comparison groups). Three expert annotators independently assessed each pair across four dimensions: Clarity, Novelty, Feasibility, and Relevance. We then computed the agreement rate between human judgments and LLM judgments. The results demonstrate strong alignment between LLM and human evaluators. Overall judgment agreement reached 90.0\% (108/120). Among individual dimensions, Clarity achieved the highest agreement rate of 90.8\% (109/120), followed by Novelty at 88.3\% (106/120), Relevance at 84.2\% (101/120), and Feasibility at 83.3\% (100/120). The average agreement across all dimensions was 87.3\% (524/600). These results are consistent with prior work~\cite{qiu2025ai,starace2025paperbench,xu2025researcherbench} showing that well-prompted LLM judges can achieve agreement rates comparable to human-human agreement (typically 80-85\%), validating the effectiveness of our automated evaluation framework.

\section{Details of Implementation}
\label{appendix:training}
\subsection{Prompts for Idea Direction and Validation Evolution}
\begin{figure*}[htbp]
  \centering
\includegraphics[width=0.9\textwidth]{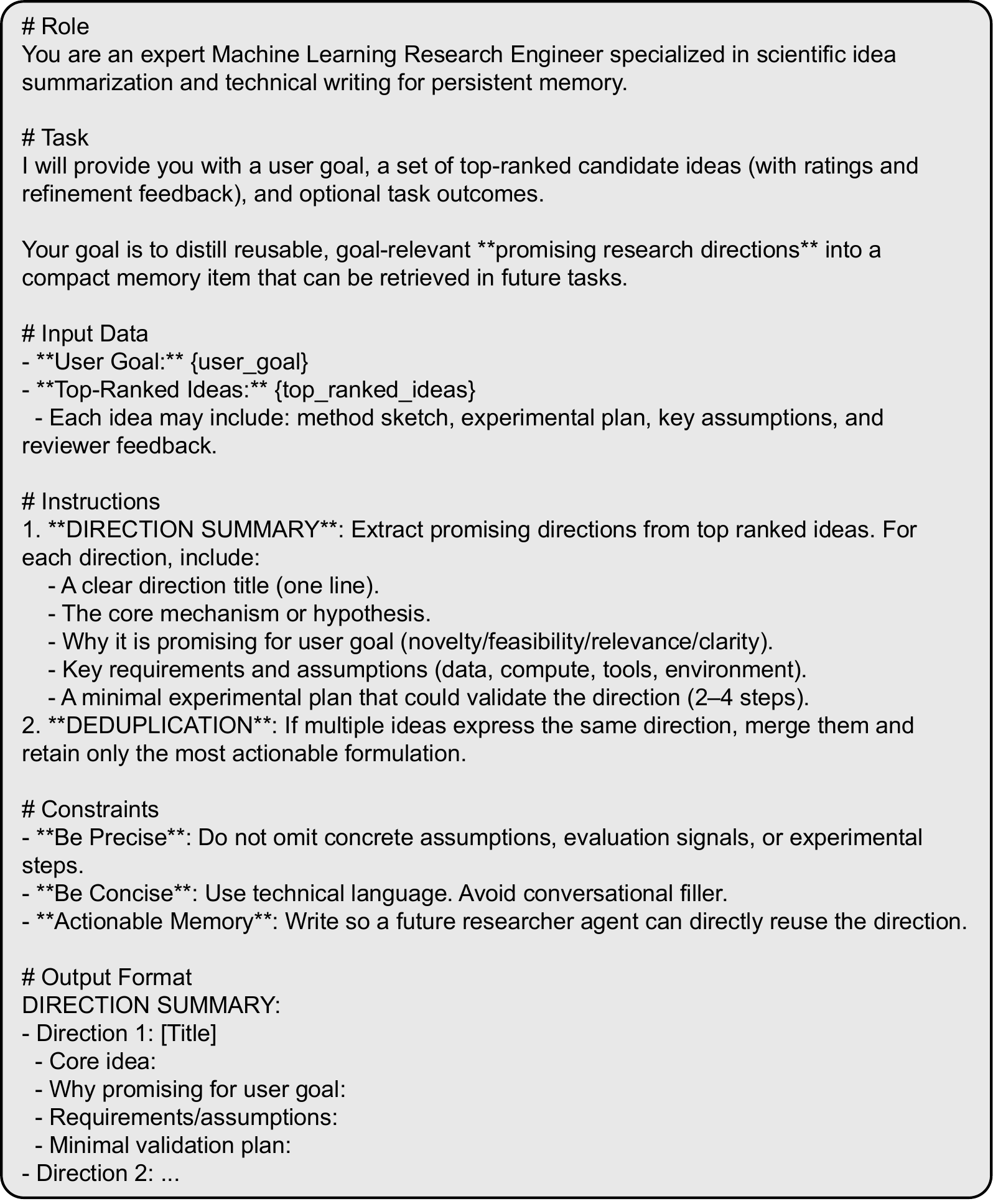}
\vspace*{-1mm}
\caption{Prompts for idea direction evolution.} 
\vspace*{-3mm}
\label{fig:ide_evo}
\end{figure*}

\begin{figure*}[htbp]
  \centering
\includegraphics[width=0.9\textwidth]{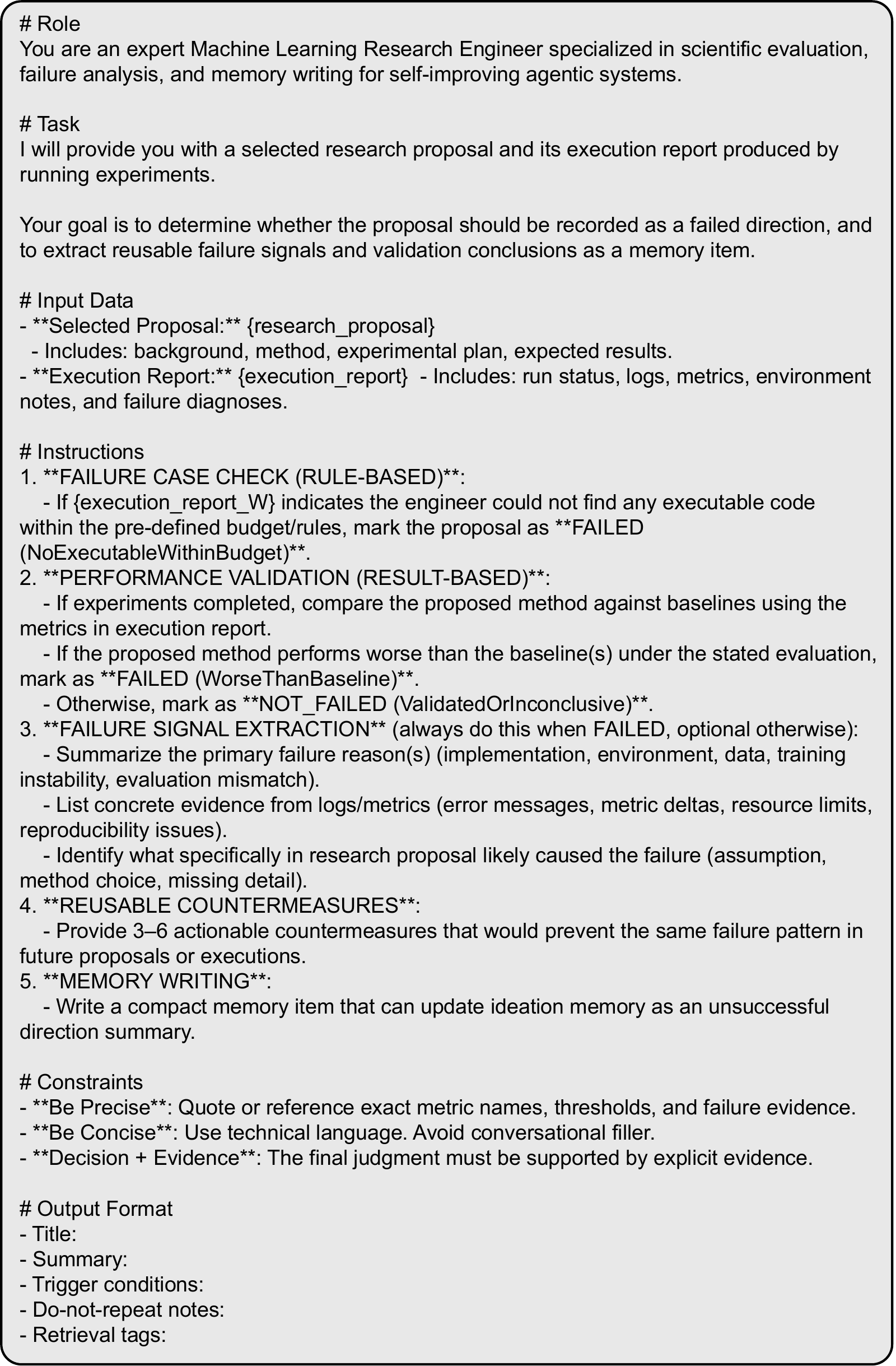}
\vspace*{-1mm}
\caption{Prompts for idea validation evolution.} 
\vspace*{-3mm}
\label{fig:ive_evo}
\end{figure*}

Details for the prompts used in $\operatorname{IDE}(\cdot)$ and $\operatorname{IVE}(\cdot)$ are provided in Figures~\ref{fig:ide_evo}, and \ref{fig:ive_evo}, respectively.

\subsection{Prompts for Experiment Strategy Evolution}
\begin{figure*}[htbp]
  \centering
\includegraphics[width=0.8\textwidth]{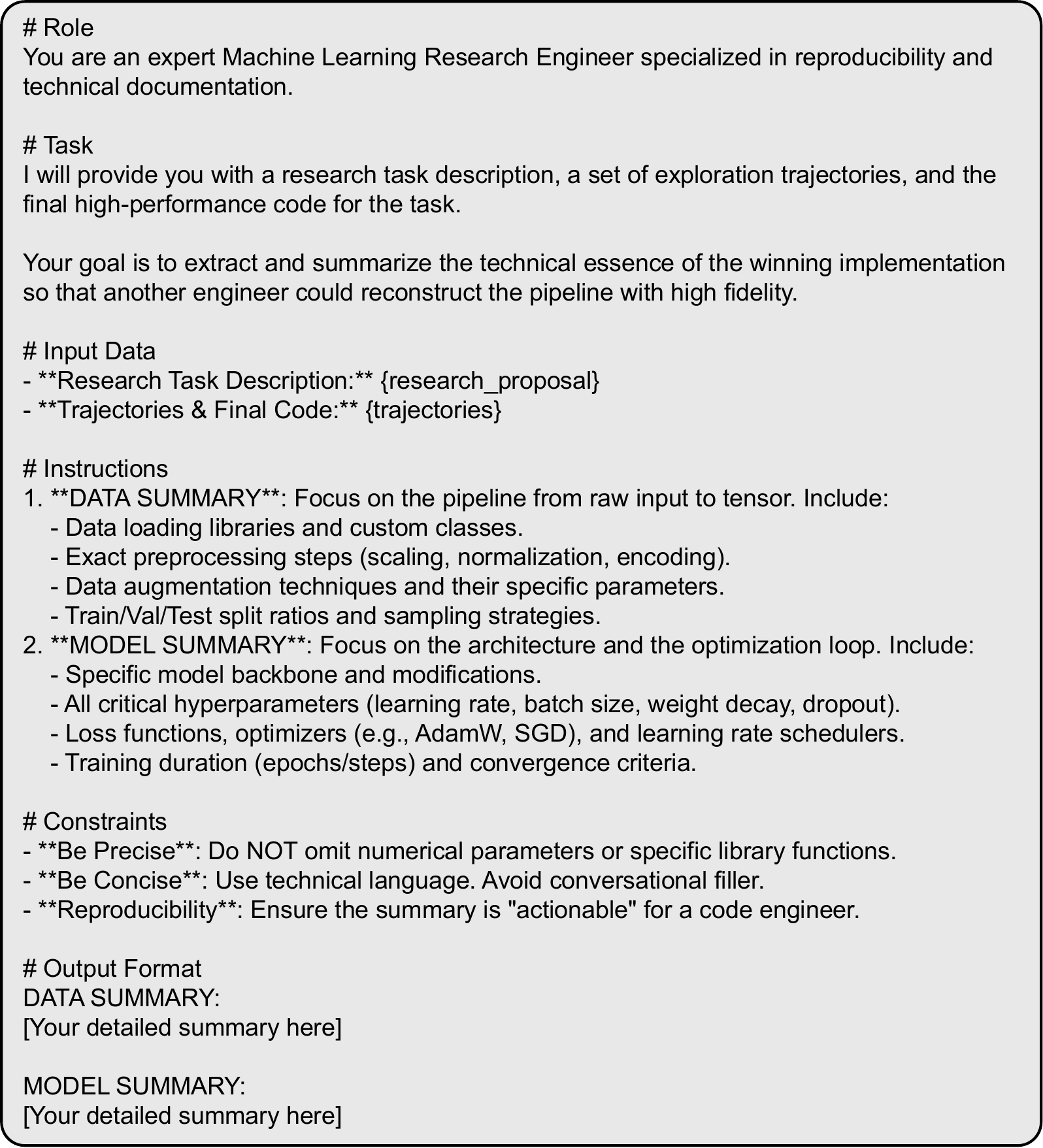}
\vspace*{-1mm}
\caption{Prompts for experiment strategy evolution.} 
\vspace*{-3mm}
\label{fig:code_evo_prompt}
\end{figure*}

Details for the prompts used in $\operatorname{ESE}(\cdot)$ is provided in Figures~\ref{fig:code_evo_prompt}.

\section{Details of Case Study}
\label{appendix:case_study_detail}

\begin{table*}[h!]
\centering
\caption{End-to-end scientific discovery performance at ICAIS 2025 (AI Scientist Track).}
\label{tab:icais_results}
\resizebox{0.7\textwidth}{!}{%
\begin{tabular}{@{} p{7.5cm} l @{}}
\toprule
\textbf{Title} & \textbf{Review Results} \\
\midrule
\href{https://airaxiv.com/papers/view/2510.0018/}{Adaptive Evidential Meta-Learning with Hyper-Conditioned Priors for Calibrated ECG Personalisation} & Best Paper Award \\
\addlinespace
\href{https://airaxiv.com/papers/view/2510.0020/}{Hierarchical Change Signature Analysis: A Framework for Online Discrimination of Incipient Faults and Benign Drifts in Industrial Time Series} & AI Reviewer's Appraisal Award \\
\addlinespace
\href{https://airaxiv.com/papers/view/2510.0023/}{Robust Zero-Shot NER for Crises via Iterative Knowledge Distillation and Confidence-Gated Induction} & Accepted \\
\addlinespace
\href{https://airaxiv.com/papers/view/2510.0022/}{Adaptive Log Anomaly Detection through Data--Centric Drift Characterization and Policy-Driven Lifelong Learning} & Accepted \\
\addlinespace
\href{https://airaxiv.com/papers/view/2510.0021/}{ConFIT: A Robust Knowledge-Guided Contrastive Framework for Financial Extraction} & Accepted \\
\addlinespace
\href{https://airaxiv.com/papers/view/2510.0019/}{Hierarchical Adaptive Normalization: A Placement-Conditioned Cascade for Robust Wearable Activity Recognition} & Accepted \\
\bottomrule
\end{tabular}%
}
\end{table*}

Table~\ref{tab:icais_results} summarizes the six accepted EvoScientist-generated papers at ICAIS 2025~\citep{icais2025} (AI Scientist Track), including links, outcomes, and condensed meta-review signals. In this section, we keep that global overview and focus the deep analysis on two representative cases: the \textit{Best Paper Award} paper and a second accepted paper with detailed reviewer diagnostics (Figure~\ref{fig:Best_Paper_Award} and Figure~\ref{fig:Appraisal_Award}).

\subsection{Best Paper Award Case: Adaptive Evidential Meta-Learning}

\begin{figure*}[htbp]
  \centering
\IfFileExists{figures/Best_Paper_Award.pdf}{
\includegraphics[width=0.9\textwidth]{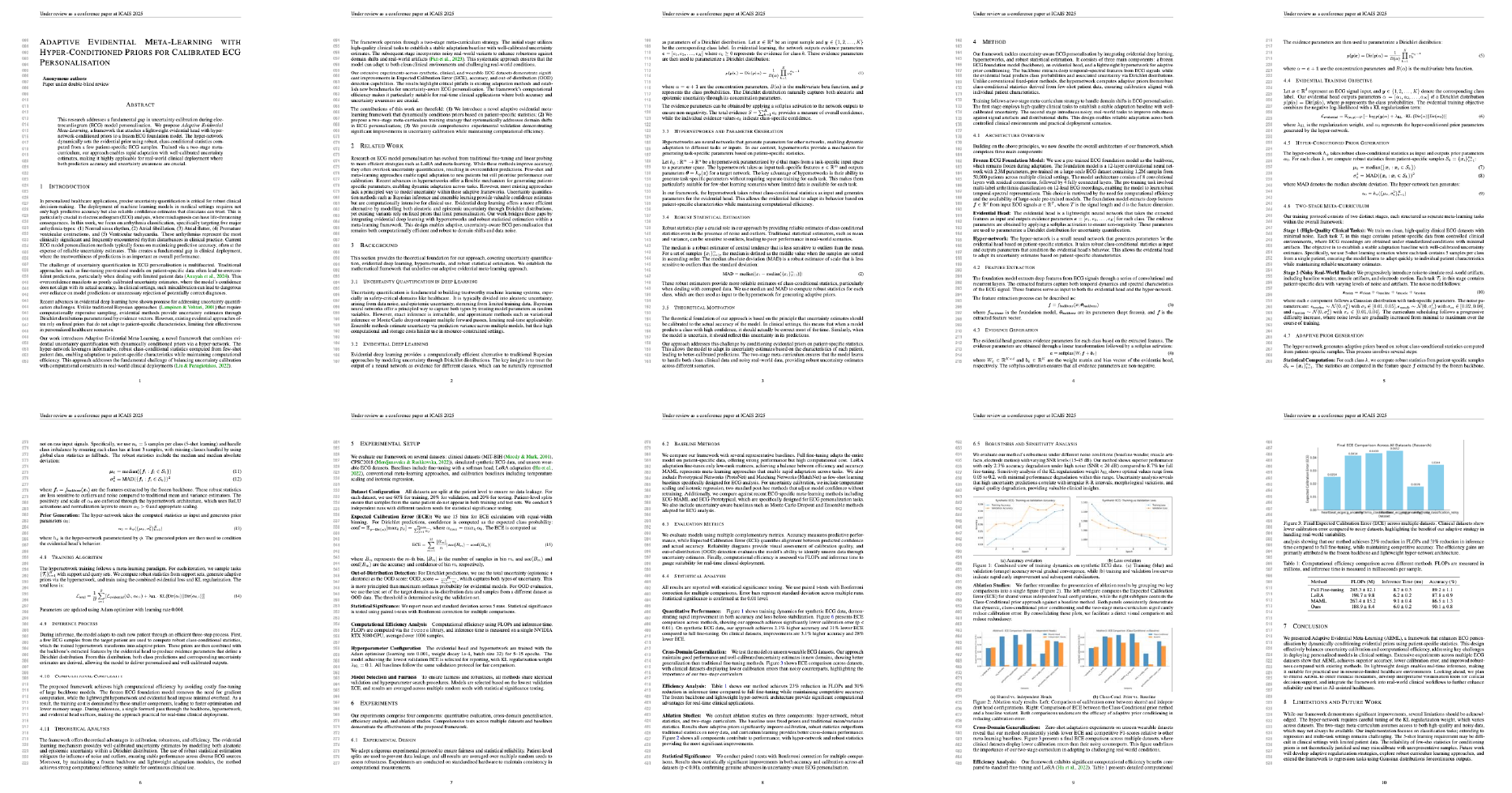}
}{
\fbox{\parbox[c][0.28\textheight][c]{0.88\textwidth}{\centering
Reserved slot for \texttt{figures/Best\_Paper\_Award.pdf}}}
}
\caption{Review evidence for \textit{Adaptive Evidential Meta-Learning with Hyper-Conditioned Priors for Calibrated ECG Personalisation} (Best Paper Award, AI Scientist Track). Original meta-review and decision page: \href{https://airaxiv.com/papers/preview/2510.0018/}{Airaxiv link}.}
\label{fig:Best_Paper_Award}
\end{figure*}

\header{\textit{Design callback}}
This case highlights the effectiveness of \textbf{idea direction evolution} within EvoScientist. As shown in Fig.~\ref{fig:Best_Paper_Award}, the researcher agent repeatedly retrieved direction-level insights from \textbf{idea memory}, enabling iterative refinement of a clinically meaningful problem formulation that balances personalization and uncertainty calibration. Reviewers emphasized the coherence and deployability of the proposed architecture, which aligns with EvoScientist's ability to reuse high-level design patterns distilled from prior outcomes rather than relying on isolated ideation.

\header{\textit{Outcome signal}}
Meta-review feedback indicates that the paper's primary strengths lie in the validity of its core contribution and its balance between methodological novelty and engineering practicality. At the same time, reviewer requests focused on clearer formalization, metric specification, and reproducibility details. This pattern is consistent with EvoScientist's design emphasis on generating empirically grounded and testable research artifacts, while leaving deeper theoretical formalization and documentation refinement as natural handover points for subsequent human-led iteration.

\subsection{AI Reviewer's Appraisal Award Case: Hierarchical Change Signature Analysis}

\begin{figure*}[htbp]
  \centering
\IfFileExists{figures/Appraisal_Award.pdf}{
\includegraphics[width=0.9\textwidth]{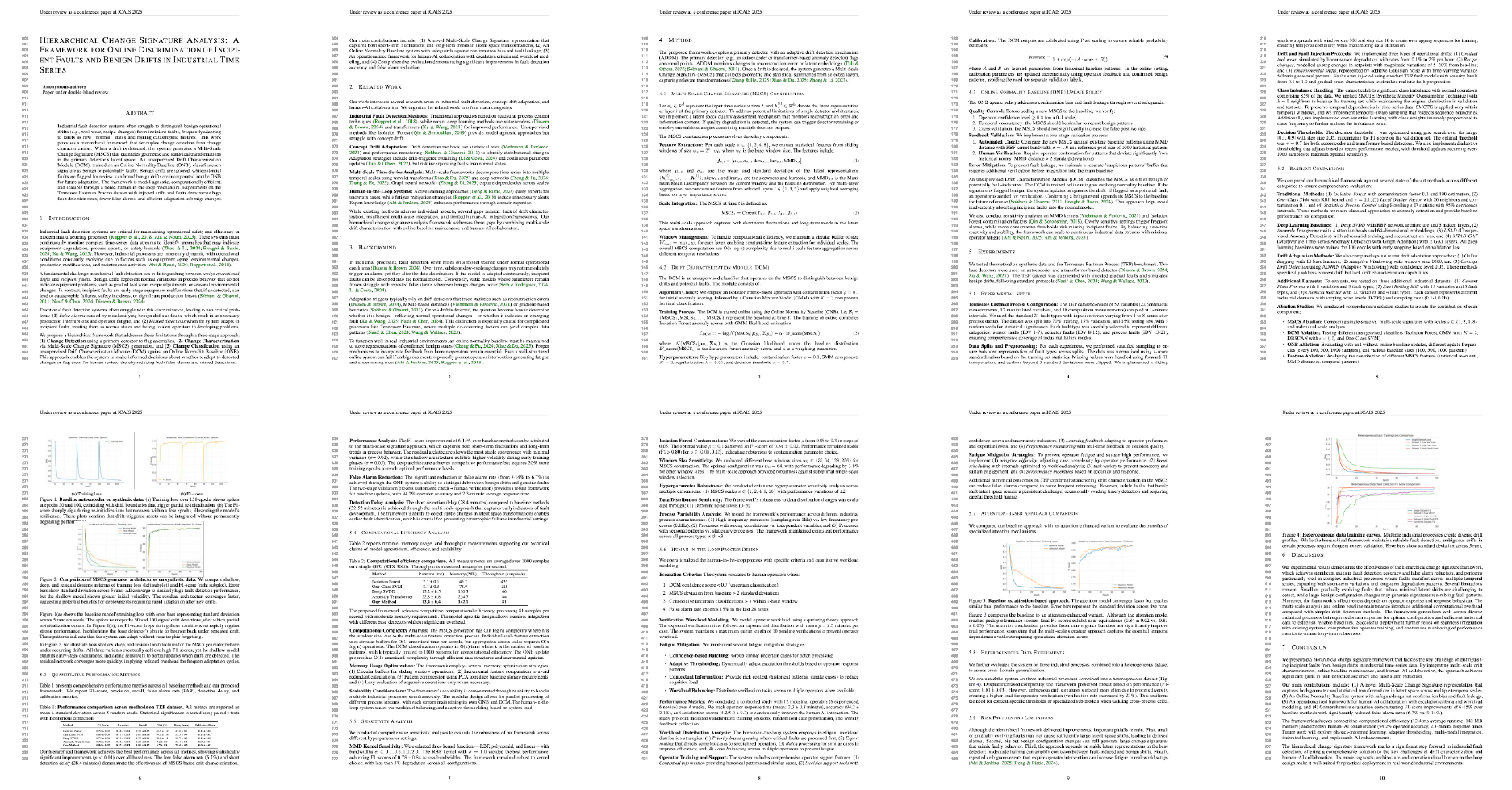}
}{
\fbox{\parbox[c][0.28\textheight][c]{0.88\textwidth}{\centering
Reserved slot for \texttt{figures/Appraisal\_Award.pdf}}}
}
\caption{Review evidence for \textit{Hierarchical Adaptive Normalization: A Placement-Conditioned Cascade for Robust Wearable Activity Recognition} (AI Reviewer's Appraisal Award, AI Scientist Track). Original meta-review and decision page: \href{https://airaxiv.com/papers/preview/2510.0019/}{Airaxiv link}.}
\label{fig:Appraisal_Award}
\end{figure*}

\header{\textit{Design callback}}
This case illustrates the role of \textbf{experiment memory} in stabilizing complex experimental pipelines. 
As shown in Fig.~\ref{fig:Appraisal_Award}, execution failures and configuration-level issues encountered during early experimentation were summarized and reused, enabling the engineer agent to converge toward a robust, deployment-oriented implementation. Reviewer praise for comprehensive empirical coverage and low runtime overhead is consistent with EvoScientist's capacity to accumulate and reuse execution-level lessons rather than repeatedly rediscovering them.

\header{\textit{Outcome signal}}
Reviewer feedback also identified several internal-consistency and protocol-level issues, including ambiguities in stability gating, metric reporting, and baseline fairness. These critiques underscore the importance of strict consistency auditing and reproducibility-complete reporting in end-to-end scientific discovery. In this sense, the appraisal outcome reflects both the strengths and current limits of EvoScientist: it is effective at producing empirically strong and practically relevant systems, while rigorous protocol alignment and documentation remain critical areas for further refinement.

Taken together, these two cases suggest that EvoScientist's evolution mechanisms can generate high-value ideas and robust empirical pipelines that align with expert evaluation criteria. At the same time, they underscore that sustained reviewer confidence depends on internal consistency, protocol-faithful baseline design, and reproducibility-complete reporting, pointing to natural directions for further system-level refinement. 

\end{document}